\documentclass[10pt,twocolumn,letterpaper]{article}

\usepackage{cvpr}              
\definecolor{cvprblue}{rgb}{0.21,0.49,0.74}
\usepackage[pagebackref,breaklinks,colorlinks,allcolors=cvprblue]{hyperref}

\usepackage{amsmath}
\usepackage{multirow}

\usepackage{amssymb}
\usepackage{pifont}
\usepackage{makecell}

\usepackage{algorithm}
\usepackage{algorithmicx}
\usepackage{algpseudocode}

\usepackage{amsmath, amssymb}   
\usepackage{graphicx}          
\usepackage{algorithm}         
\usepackage{algpseudocode}     
\usepackage{xcolor}            
\usepackage{lipsum}            

\algrenewcommand\algorithmiccomment[1]{\hfill\textcolor{blue}{\scalebox{0.9}{$\triangleright$}} \textcolor{blue}{\textit{#1}}}

\def\paperID{11975} 
\def\confName{CVPR}
\def\confYear{2026}

\title{Pluggable Pruning with Contiguous Layer Distillation for Diffusion Transformers}

\author{Jian Ma$^{1}$\\
OPPO AI Center\\
{\tt\small majian2@oppo.com}
\and
Qirong Peng$^{1}$\\
OPPO AI Center\\
{\tt\small pengjirong@oppo.com}
\and
Xujie Zhu$^{1,2}$\\
Sun Yat-sen University\\
{\tt\small zhuxj6@mail2.sysu.edu.cn}
\and
Peixing Xie$^{2}$\\
The Chinese University of Hong Kong \\
{\tt\small 1155254853@link.cuhk.edu.hk}
\and
Chen Chen\\
OPPO AI Center\\
{\tt\small chenchen4@oppo.com}
\and
Haonan Lu\\
OPPO AI Center\\
{\tt\small luhaonan@oppo.com}
}

\begin{document}
\maketitle
\footnotetext[1]{Co-first authors.}
\footnotetext[2]{The author did his work during internship at OPPO AI Center.}
\begin{abstract}
Diffusion Transformers (DiTs) have shown exceptional performance in image generation, yet their large parameter counts incur high computational costs, impeding deployment in resource-constrained settings. To address this, we propose Pluggable Pruning with Contiguous Layer Distillation (PPCL), a flexible structured pruning framework specifically designed for DiT architectures. First, we identify redundant layer intervals through a linear probing mechanism combined with the first-order differential trend analysis of similarity metrics. Subsequently, we propose a plug-and-play teacher-student alternating distillation scheme tailored to integrate depth-wise and width-wise pruning within a single training phase. This distillation framework enables flexible knowledge transfer across diverse pruning ratios, eliminating the need for per-configuration retraining. Extensive experiments on multiple Multi-Modal Diffusion Transformer architecture models demonstrate that PPCL achieves a 50\% reduction in parameter count compared to the full model, with less than 3\% degradation in key objective metrics. Notably, our method maintains high-quality image generation capabilities while achieving higher compression ratios, rendering it well-suited for resource-constrained environments.  The open-source code, checkpoints for PPCL can be found at the following link: https://github.com/OPPO-Mente-Lab/Qwen-Image-Pruning.
\end{abstract}

\section{Introduction}

\begin{figure*}[t]
\centering
\includegraphics[width=1\textwidth]{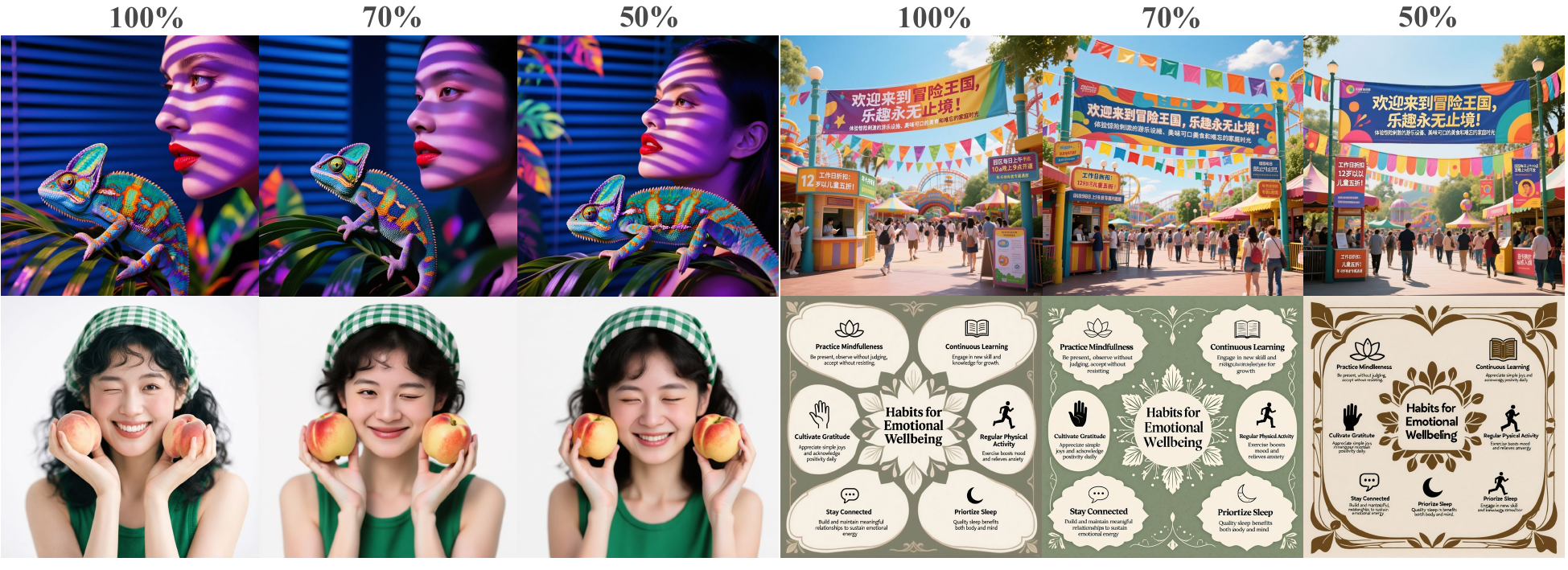}
\caption{Visual comparison of Qwen-Image and its progressive pruned variants. Columns 1 and 4 show the original 20B-parameter Qwen-Image; Columns 2 and 5 show the 70\% parameter variant. Columns 3 and 6 show the 50\% variant. Results demonstrate that pruned variants retain generation quality comparable to the original model in color rendering, fine-grained text details, and facial feature synthesis.
}\label{overall}
\end{figure*}

\begin{figure}[t]
\centering
\includegraphics[width=0.5\textwidth]{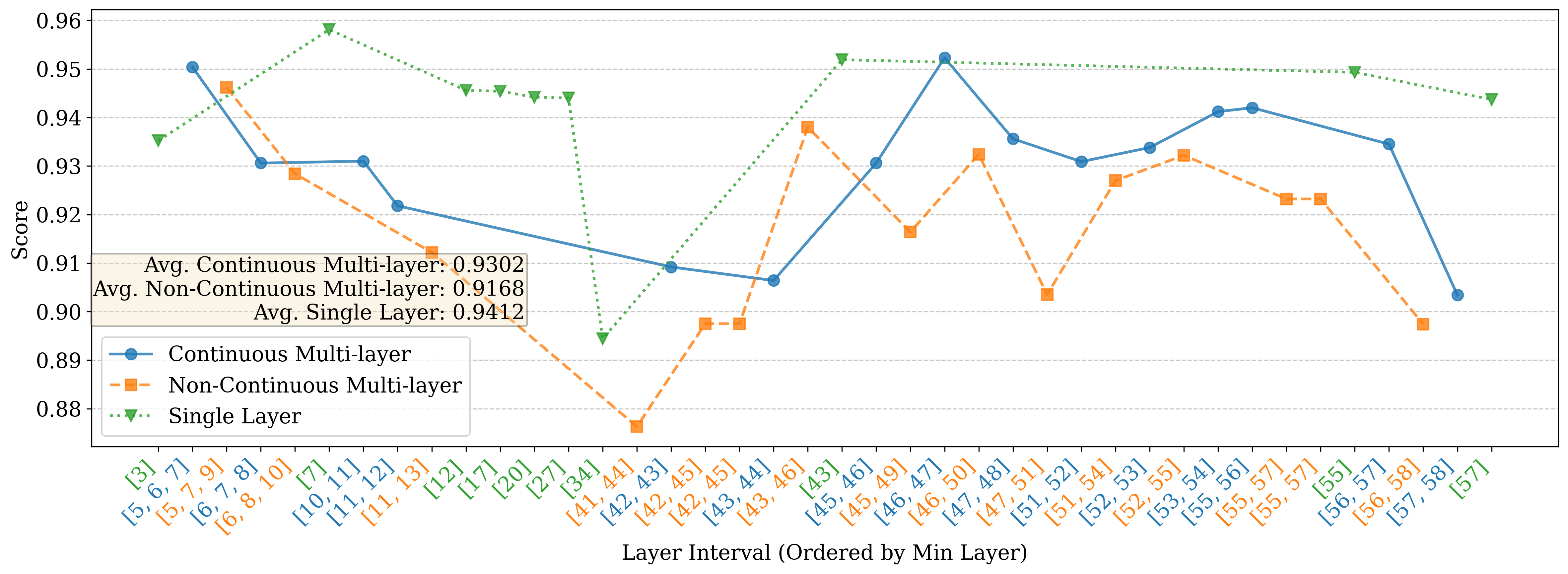}
\caption{Performance of Qwen-Image on LongText-Bench under three layer removal strategies: individual layers, contiguous layers, and non-contiguous layers. The x-axis denotes the index of the removed layer(s), and the y-axis indicates the accuracy. The pale-yellow zone in the lower-left indicates the mean accuracy for each of the three removal strategies.}\label{contiguous}
\end{figure}

DiTs have emerged as the dominant architecture for high-quality Text-to-Image (T2I) generation. Recent models including Stable Diffusion 3.5 (SD3.5)\cite{SD3.5}, FLUX.1\cite{flux} and Qwen-Image\cite{Qwen-image} significantly outperform earlier approaches such as SDXL\cite{SDXL}, SD1.5\cite{SD}, and DALL·E 2\cite{Dalle2} in both image fidelity and text-image alignment. However, this leap in performance comes at a steep computational cost: state-of-the-art DiTs typically contain 8–20 billion parameters, demanding substantial resources during both training and inference. This scale severely restricts deployability in resource-constrained scenarios, underscoring the need for efficient model compression.

While existing work on diffusion model optimization has largely focused on accelerating sampling or applying general-purpose compression methods, structured pruning\cite{fang2023structuralpruningdiffusionmodels,EcoDiff,yu2025comp,dong2025tinysr} has emerged as a particularly promising direction due to its hardware-friendliness and seamless compatibility with other efficiency-enhancing techniques, such as INT4/INT8 quantization\cite{shang2023post,li2023q,so2023temporal}, knowledge distillation\cite{huang2023knowledge,sauer2024adversarial,ma2025x2iseamlessintegrationmultimodal,meng2023distillation,salimans2022progressive}, and neural architecture search\cite{asthana2024multi,tang2023lightweight,mecharbat2024dfusenas}. Despite these advantages, current structured pruning approaches for diffusion models suffer from three critical limitations: (1) Most methods lack generalizability to various mainstream Multi-Modal Diffusion Transformers (MMDiTs)\cite{SD3.5}. (2) They offer limited flexibility in layer pruning and provide little support for plug-and-play configurations. (3) There remains insufficient understanding of inter-layer dependencies in deep diffusion models.

This work focuses on the MMDiT architecture. Recent studies\cite{ostris,flux1-lite} have shown that random layer removal from FLUX.1 causes marginal performance degradation. To explore layer redundancy patterns, we conduct comprehensive experiments on Qwen-Image, a 60-layer MMDiT with 20B parameters, where 1, 2, or 3 layers are removed in contiguous and non-contiguous configurations, and analyze their impact on generation quality. As illustrated in Fig.\ref{contiguous}, our experiments reveal two observations: (1) Random layer removal has minimal effect on generation quality, indicating considerable layer redundancy. (2) Contiguous removal consistently outperforms non-contiguous removal, suggesting that redundancy exhibits strong depth-wise continuity. The second observation gives rise to a core question: Is there a strategy that can maximize the identification of subsets of contiguous redundant layers? Our key insight is that representation evolution across layers in the teacher model is not uniform, but rather progresses in distinct phases. Within each phase, layer activations exhibit smooth transitions, permitting compression without notable performance loss. Furthermore, when the input-output mapping of a layer can be approximated by a linear function, its transformation lies close to a linear subspace and is functionally redundant to adjacent layers\cite{alain2016understanding, molchanov2016pruning}. Based on the above, we implement a strategy for detecting redundant layers. We establish substitutability as the core criterion for measuring layer redundancy. Specifically, we construct a linear probe for each layer of the teacher model and train the probe using alignment loss to approximate the input-output mapping of the layer. Subsequently, by analyzing the Centered Kernel Alignment (CKA)\cite{CKA} of layer activations and the concavity patterns of their first-order differences on a calibration dataset, we maximize the identification of subsets of contiguous redundant layers prior to distillation.

In conventional sequential distillation, errors from early-layer compression propagate and compound through the network, leading to semantically misaligned representations in the student model. Additionally, intermediate-layer distortions cause the distillation objective to diverge from the teacher’s conditional distribution. To address these issues, we propose a non-sequential inter-layer distillation: Student layers directly take outputs of the teacher’s immediately preceding depth as inputs, aligning the original representation flow. This design breaks the error propagation chain, enables independent optimization of pruned modules, and realizes a modular training paradigm. Unlike prior non-pluggable pruning/distillation methods, our approach allows arbitrary activation or bypassing of specific layers at inference. This supports flexible trade-offs between inference speed and generation fidelity without retraining.


Though depth-wise pruning reduces model depth, MMDiT retains significant width-wise redundancy. Analyzing its multimodal nature reveals two key sources: (1) Stream-level redundancy: Text streams show higher token similarity and lower inter-layer variability, enabling heavy compression. (2) Feed-Forward Network (FFN) redundancy: FFNs in both text and image streams are significantly over-parameterized, with linear projectors capable of effectively approximating their function. Thus, we introduce width-wise pruning, which replaces text streams and FFNs with compact lightweight linear projectors. This dual-axis compression yields significant size reduction, preserves generation quality and text-image alignment, and further mitigates distillation objective deviation from the teacher-induced distribution.

Experiments on multiple T2I models demonstrate that our method consistently outperforms existing compression approaches. Notably, on FLUX.1 and Qwen-Image, it reduces model size to 30–50\% of the original, achieves 1.3–1.8× inference speedup, and cuts GPU memory consumption by over 30\% while preserving visual fidelity and semantic alignment comparable to the teacher model. Our contributions are summarized as follows:
\begin{itemize}
    \item We investigate layer redundancy patterns in MMDiT, uncovering the depth-wise continuity of redundant layers.
    \item We implement a lightweight linear probe-based redundancy detection strategy, integrating the linear approximability of input-output mappings with CKA-based representation trajectory analysis to maximize the identification of subsets of redundant layer sets.
    \item We introduce a depth-wise pruning framework that suppresses error propagation and enables dynamic inference-time pruning without fine-tuning. We further integrate width-wise pruning to achieve a more compact model.
    \item We validate our approach across multiple MMDiT models, demonstrating significant efficiency gains with minimal performance degradation.
\end{itemize}
\section{Related Work}

\subsection{Text-to-image Diffusion Models}

T2I models have rapidly evolved from early U-Net architectures to MMDiT, now used in advanced models like SD3.5, FLUX.1, and Qwen-Image. MMDiT improves semantic alignment and visual quality by jointly modeling image latent representations and text embeddings. However, the parameter count has increased from 2.6B in SDXL to 20B in Qwen-Image, hindering deployment on real-time systems. This highlights the necessity for compression methods that significantly reduce computational costs while preserving generation quality.

\subsection{Efficient Model Compression with Pruning}

To mitigate computational burdens in large models, extensive research has focused on model compression via pruning and knowledge distillation, including structured\cite{muralidharan2024compact, taghibakhshi2025efficient, Shearedllama, Llm-pruner} and unstructured pruning\cite{nordstrom2022unstructured,kurtic2022optimal}. SparseGPT\cite{SparseGPT} and Wanda\cite{Wanda} employ sparse regression and weight-activation products for training-free compression, whereas SlimGPT\cite{SlimGPT} presents a low-cost and fast structured pruning method through Optimal Brain Surgeon\cite{hassibi1993optimal} framework.


For T2I diffusion models, pruning efforts have primarily aimed to reduce computational costs while avoiding costly retraining. Extensive studies focus on U-Net architectures, with training-free\cite{EcoDiff,guo2025mosaicdiff} and training-based\cite{LD-Pruner,zhang2024laptop,ganjdanesh2024not} approaches. SnapFusion\cite{Snapfusion}, BK-SDM\cite{Bk-sdm}, and KOALA\cite{Koala} enable pruning of redundant residual blocks or attention modules through distillation-based sensitivity analysis.


The emergence of DiTs has driven the development of Transformer-specific compression techniques\cite{FastFLUX, DiP-GO}. TinyFusion\cite{Tinyfusion} employs differentiable layer sampling for high-recoverability shallow subnetworks. HierarchicalPrune\cite{HierarchicalPrune} compresses the model via Hierarchical Position Pruning (HPP) and Positional Weight Preservation (PWP). Dense2MoE\cite{Dense2MoE} reduces activation costs while maintaining overall model capacity by replacing FFNs with Mixture of Experts\cite{shazeer2017outrageously}. Recent distillation-based approaches yield compact and high-performance T2I models. FLUX.1 Lite\cite{flux1-lite} (8B) achieves a 20\% inference speedup and 7GB memory reduction while preserving visual fidelity. Chroma1-HD\cite{Chroma1-HD} (8.9B) improves controllability through optimized timestep encoding and MMDiT masking. These advances demonstrate that structured pruning combined with knowledge distillation enables deployable, high-fidelity T2I systems.


\section{Method}

\begin{figure}[t]
\centering
\includegraphics[width=0.5\textwidth]{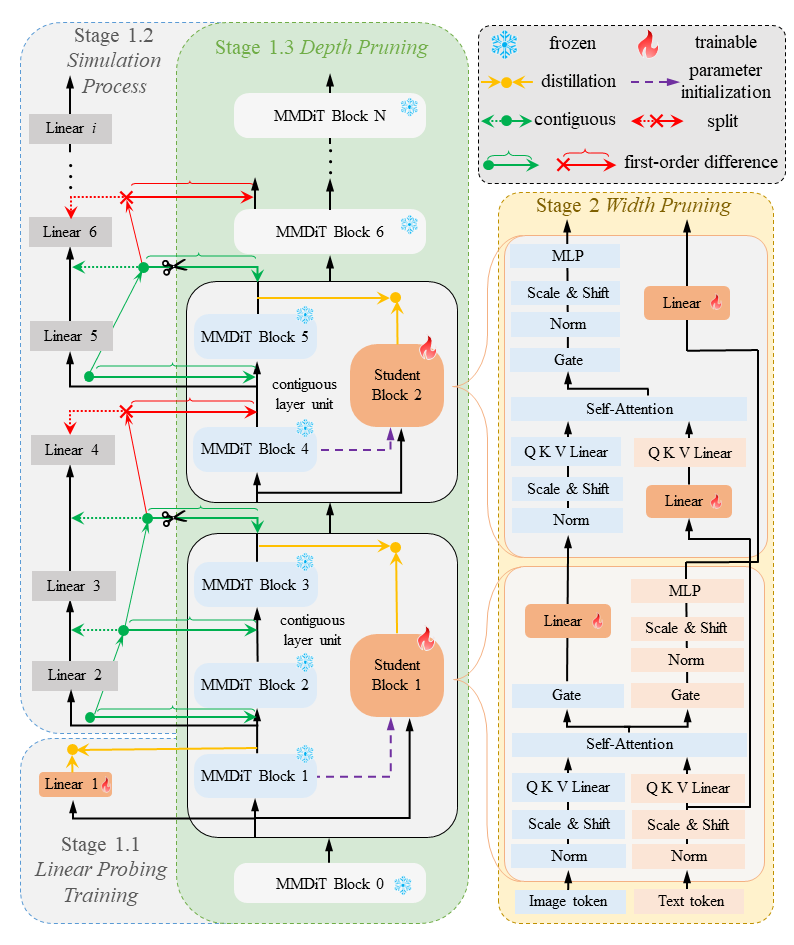}
\caption{(a) Depth-wise pruning: Stage 1.1 performs linear probing training for each MMDiT block. Stage 1.2 simulates pruning training to assess the continuity between adjacent MMDiT blocks by tracking the first-order difference of CKA between each block outputs and its corresponding linear probe outputs. A decreasing first-order difference indicates a contiguous layer, while a sudden increase suggests a break. The length represents the value of the first-order difference. Stage 1.3 conducts feature distillation, with the inputs to the student model taken from the same contiguous layer unit. (b) Width-wise pruning: We prune both stream-level and FFN redundancy in MMDiT. 
}\label{model}
\end{figure}

Our overall framework and procedure, as illustrated in Fig.\ref{model} and Algorithm \ref{PPCL_algorithm}, consist of two stages: Depth-wise pruning and width-wise pruning. The first stage is redundant intervals detection (Sec.\ref{3.2}, gray region in Fig.\ref{model}), a process involving training followed by simulation to identify contiguous redundant layers. We then perform the actual depth-wise pruning (Sec.\ref{3.3}, light green region in Fig.\ref{model}). The second stage is width-wise pruning (Sec.\ref{3.4}, rightmost part of Fig.\ref{model}). Finally, we perform a brief fine-tuning.

\begin{algorithm}[t]
\footnotesize
\caption{PPCL for Diffusion Transformers}
\label{PPCL_algorithm}
\begin{algorithmic}[1]

    \Statex \textbf{Input:}
    \Statex \hspace{\algorithmicindent} Teacher model $\mathcal{T}$ with $M$ MMDiT blocks, parameters $\theta_\mathcal{T}$ and parameter count $P_\mathcal{T}$, training set $D$ and calibration set $X$, least-squares-initialized linear probes $\mathcal{L}_p = \{ l_i \}_{i=1}^M$, sets of width-wise redundant layers $\mathcal{R}_{txt}$ and $\mathcal{R}_{ffn}$.
    \Statex \textbf{Output:}
    \Statex \hspace{\algorithmicindent} Student model $\mathcal{S}$ with parameters $\theta_\mathcal{S}$ and parameter count $P_\mathcal{S}$

    \medskip

    \Statex \textbf{Stage 1.1: Linear probing training (Sec.\ref{3.2})}
    \State $\mathcal{L}_{fit}(i) \leftarrow \text{CalculateFitLoss}(l_i, \mathcal{T}, D, i),\forall i\in[1,M]$(Eq.\ref{fit loss})
    \State $l_i^{\prime} \leftarrow \text{UpdateParams}(l_i, \mathcal{L}_{fit}(i)),\forall i\in[1,M] $
    \State $\mathcal{L}_p^{\prime} \leftarrow \{ l_i^{\prime} \}_{i=1}^M$
    
    \medskip

    \Statex \textbf{Stage 1.2: Simulation process (Sec.\ref{3.2})}
    \State $\mathcal{I} \gets \emptyset$, $u \gets 1$
    \State \textbf{while} $u \leq M$ \textbf{do}
        \State \hspace{\algorithmicindent} ${\Delta}(u,k) \gets $
        \Statex \hspace{\algorithmicindent} \hspace{\algorithmicindent} $\text{CalculateDelta}(\mathcal{T}, \mathcal{L}_p^{\prime}, X, u, k),\forall k\in[u+1,M]$(Eq.\ref{delta})
        \State \hspace{\algorithmicindent} \textbf{for} $k \in \{u+2, \ldots, M\}$ \textbf{do} (Eq.\ref{detect})
            \State \hspace{\algorithmicindent} \hspace{\algorithmicindent} \textbf{if} $\Delta(u,k) > \Delta(u,k-1)$
                \State \hspace{\algorithmicindent} \hspace{\algorithmicindent} \hspace{\algorithmicindent} $v \gets k-1$
                \State \hspace{\algorithmicindent} \hspace{\algorithmicindent} \hspace{\algorithmicindent} \textbf{break}
            \State \hspace{\algorithmicindent} \hspace{\algorithmicindent} \textbf{end if}
        \State \hspace{\algorithmicindent} \textbf{end for}
        \State \hspace{\algorithmicindent} $\mathcal{I} \gets \mathcal{I} \cup \{[u, v]\}$
        \State \hspace{\algorithmicindent} $u \gets v + 1$
    \State \textbf{end while}

    \medskip

    \Statex \textbf{Stage 1.3: Depth-wise pruning (Sec.\ref{3.3})}
    \State $\mathcal{L}_{depth} \gets$ $0$
    \State $\mathcal{S}_{init}, \theta_{init}, P_{init} \gets \text{Initial}(\mathcal{T},\theta_\mathcal{T}, P_\mathcal{T})$
    \State \textbf{for} each interval $[u, v] \in \mathcal{I}$ \textbf{do}
        \State \hspace{\algorithmicindent} $T_{u-1}^D \gets T_{1:u-1}(D)$ 
        \State \hspace{\algorithmicindent} $\mathcal{L}_{depth}^{[u,v]} \gets \text{CalculateDistillLoss}(T, S_{init}, T_{u-1}^D, u, v)$(Eq.\ref{depth total})
        \State \hspace{\algorithmicindent} $\mathcal{L}_{depth} \leftarrow \mathcal{L}_{depth} + \mathcal{L}_{depth}^{[u,v]}$
    \State \textbf{end for}
    \State $\mathcal{S}_{depth}, \theta_{depth}, P_{depth} \leftarrow$
    \Statex \hspace{\algorithmicindent} $\text{UpdateParams}(\mathcal{S}_{init}, \theta_{init}, P_{init},\mathcal{L}_{depth})$

    \medskip

    \Statex \textbf{Stage 2: Width-wise pruning (Sec.\ref{3.4})}
    \State $\mathcal{L}_{width} \gets$ $0$
    \State $\mathcal{S}_{width}, \theta_{width}, P_{width} \gets$ 
    \Statex \hspace{\algorithmicindent} $\text{ReplaceWithLinear}(\mathcal{S}_{depth}, \theta_{depth}, P_{depth}, \mathcal{R}_{txt}, \mathcal{R}_{ffn})$
    \State \textbf{for} $j \in \mathcal{R}_{txt} \cup \mathcal{R}_{ffn}$ \textbf{do}
        \State \hspace{\algorithmicindent} $T_{j-1}^D \gets T_{1:j-1}(D)$ 
        \State \hspace{\algorithmicindent} $\mathcal{L}_{width}^{j}, \mathcal{L}_{linear}^{j} \gets $
        \Statex \hspace{\algorithmicindent} \hspace{\algorithmicindent} $\text{CalculateDistillLoss}(T, S_{width}, T_{j-1}^D, j)$(Eq.\ref{width loss}, Eq.\ref{width linear})
        \State \hspace{\algorithmicindent} $\mathcal{L}_{width} \leftarrow \mathcal{L}_{width} + (\mathcal{L}_{width}^{j}+\mathcal{L}_{linear}^{j})$
    \State \textbf{end for}
    \State $\mathcal{S}, \theta_{\mathcal{S}}, P_{\mathcal{S}} \leftarrow$
    \Statex \hspace{\algorithmicindent} $\text{UpdateParams}(\mathcal{S}_{width}, \theta_{width}, P_{width},\mathcal{L}_{width})$

    \medskip
    
    \State \Return $\mathcal{S}, \theta_{\mathcal{S}}, P_{\mathcal{S}}$ 
\end{algorithmic}
\end{algorithm}

\subsection{Motivations}

\begin{figure*}[t]
\centering
\includegraphics[width=1\textwidth]{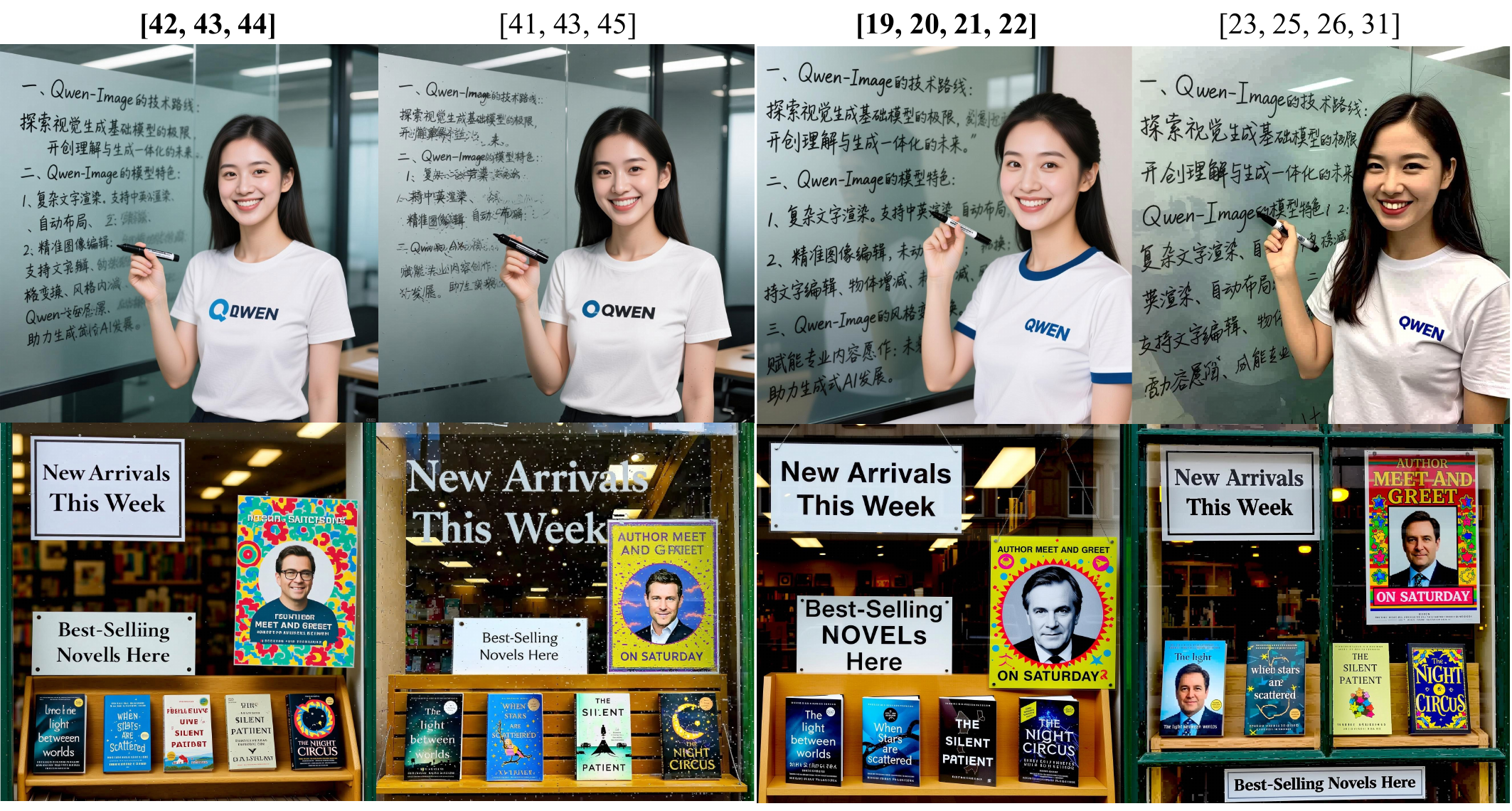}
\caption{Subjective comparison of complex text rendering in Qwen-Image when randomly removing contiguous and non-contiguous blocks.
Columns 1 and 3 show the results for contiguous layer removal, while columns 2 and 4 correspond to non-contiguous layer removal.}\label{Consecutive-NonConsecutive}
\end{figure*}

In T2I diffusion models, the early layers encode foundational semantics and global layout, while the later layers refine high-frequency details. Simultaneously, within the computational graph of a deep generative model, adjacent layers tend to process hierarchically related semantic content. For instance, during inference, one contiguous block may collaboratively construct the high-level structure of an object, while a subsequent block refines its texture details. 
As illustrated in Fig.\ref{contiguous} and Fig.\ref{Consecutive-NonConsecutive}, the overall generation quality shows minimal degradation even when one to three layers are randomly removed. This observation suggests significant architectural redundancy and functional overlap among layers, implying that redundant layers can be pruned without notable loss in visual quality. Furthermore, removing contiguous layers results in only marginal performance degradation, consistently outperforming both random and non-contiguous removal. This indicates that these layers are not functionally independent but instead operate as tightly coupled functional units. 

We further elaborate that, our definition of redundant layers is not evaluated based on the minimal loss incurred by direct layer removal, but rather on a criterion of substitutability. Subsequently, we need to identify the substitutability of consecutive layers. To efficiently detect multiple intervals of redundant consecutive layers without exhaustive search, we implement a novel strategy that leverages lightweight linear probes to model layer substitutability, requiring only a single training process. This strategy is underpinned by two technical pillars: (a) As illustrated in Stage 1.1 of Fig.\ref{model}, the input for the linear probe during training is consistent with the input to the corresponding layer module, ensuring that the modeling of each layer is independent and unaffected by other layers; (b) The superposition of a finite number of linear transformations satisfies linearity, thereby ensuring the modeling of substitutability for consecutive layers.

\subsection{Redundant Intervals Detection}
\label{3.2}

The objective of the detection is to identify a set $\mathcal{I} \!=\! \{[u_i, v_i]\}_{i=1}^n$ ($n = |\mathcal{I}|$), which is composed of multiple redundant intervals. Let the teacher model be denoted as $\mathcal{T} \!=\! \{T_1, T_2, \dots, T_M\}$, with a corresponding set of linear probes $\mathcal{L}_p \!=\! \{l_1, l_2, \dots, l_M\}$, where $M$ is the total number of layers. Since MMDiT incorporates residual connections, we equip each probe $l_i$ with a residual structure to preserve inductive bias compatibility. Specifically, for a training set $D$, we initialize $l_i$ with the optimal weight matrix $W_i^*$ obtained by solving a least-squares problem $ \min_{W_i} \left\| (\mathbf{I} + W_i) T_{i-1}^D - T_i(T_{i-1}^D) \right\|_2^2$, whose closed-form solution is given by:
\begin{equation}
\label{least-squares}
W_i^* = \big( T_i(T_{i-1}^D) - T_{i-1}^D \big) (T_{i-1}^D)^\top \big( T_{i-1}^D (T_{i-1}^D)^\top \big)^{-1},
\end{equation}
where $\mathbf{I}$ denotes the identity matrix, $T_{i-1}^D$ denotes the outputs of the $(i-1)$-th teacher layer. Subsequently, as shown in Stage 1.1 of Algorithm \ref{PPCL_algorithm}, we train each $l_i$ by minimizing the alignment loss between the teacher layer $T_i$ and the corresponding linear probe:
\begin{equation}
\label{fit loss}
\mathcal{L}_{fit}(i) = \left\| l_i(T_{i-1}^D) + T_{i-1}^D - T_i(T_{i-1}^D) \right\|_2^2.
\end{equation}

After training, we perform inference on a calibration set $X$ and record the activations of both the linear probes and the corresponding teacher layers. As shown in Stage 1.2 of Algorithm \ref{PPCL_algorithm}, we then analyze the feature evolution across layers by computing CKA between layer representations to identify candidate pruning intervals. Specifically, for a given starting layer $u$ and a variable $k$ ($k\in[u+1,M]$), let $T^{[u \to k]}$ denote the surrogate model obtained by replacing $T_{u+1}, \dots, T_k$ with their corresponding linear probes. We compute the CKA similarity between the outputs of the original teacher layer $T_u$ and the outputs of the surrogate model at layer $k$:
\begin{equation}
\label{CKA}
\mathrm{cka}(u, k) = \mathrm{CKA}\big( T_u(T_{u-1}^X),\, T_{u:k}^{[u \to k]}(T_{u-1}^X) \big),
\end{equation}
where $T_{u-1}^X$ denotes the outputs of the $(u-1)$-th teacher layer and $T_{u:k}^{[u \to k]}$ denotes layers $u$ through $k$ of $T^{[u \to k]}$. We define the first-order difference along the layer index as:
\begin{equation}
\label{delta}
\Delta(u, k) = -(\mathrm{cka}(u, k) - \mathrm{cka}(u, k-1)).
\end{equation}
The endpoint $v$ corresponding to $u$ is determined by the trend of $\Delta(u, k)$. A redundant interval $[u, v]$ is characterized by a local minimum in $\Delta(u, k)$, that is, $\Delta(u, k)$ first decreases and then increases. This inflection point indicates that the rate of decline in representational similarity slows down and eventually reverses, signaling that additional layers begin to contribute meaningfully again and thus marking the end of a redundant interval. Formally, we define:
\begin{equation}
\label{detect}
v = \min \left\{ k-1 \in [u+2, M] \mid \Delta(u, k) > \Delta(u, k-1) \right\}.
\end{equation}

Iteratively applying this procedure yields multiple redundant layer intervals that collectively form the set $\mathcal{I}$. This detection strategy identifies regions with stable representational similarity, which are amenable to contiguous-layer pruning, as well as sharp boundaries with abrupt fidelity changes that serve as natural boundaries. Notably, this approach aligns with our subsequent distillation framework.

\subsection{Depth-wise Pruning of Contiguous Layers}
\label{3.3}

In traditional sequential distillation, compression errors in the early layers propagate and compound, yielding student representations that are semantically misaligned with those of the teacher. To address this issue, we propose a non-sequential distillation scheme.

In Stage 1.3 of Algorithm \ref{PPCL_algorithm}, for a given interval $[u, v] \in \mathcal{I}$, we initialize the student layer $S_{init}^u$ with the weights of teacher layer $T_u$. The distillation loss is defined as:
\begin{equation}
\label{depth loss}
\mathcal{L}_{depth}^{[u,v]} = \left\| \mathrm{Norm}\left(S_{init}^u(T_{u-1}^D)\right) - \mathrm{Norm}\left(T_v^D\right) \right\|_2^2,
\end{equation}
where $T_{u-1}^D$ and $T_v^D$ denote the outputs of the $(u-1)$-th and $v$-th teacher layers, respectively. This aligns the outputs of the student model with the representation distribution at the $v$-th teacher layer, mitigating the distributional shifts. $\mathrm{Norm}(\cdot)$ represents L2 normalization along the feature dimension, which emphasizes directional alignment and stabilizes training. The total loss is given by:
\begin{equation}
\label{depth total}
\mathcal{L}_{depth} = \sum_{i=1}^{n} \mathcal{L}_{depth}^{[u_i,v_i]}.
\end{equation}
This design isolates the optimization of each interval, thereby preventing error accumulation and enabling modular training. Furthermore, it supports dynamic adjustment of the pruning ratio by invoking original teacher layers as needed, achieving a flexible trade-off between inference speed and representation fidelity.

\subsection{Width-wise Pruning}
\label{3.4}

\begin{figure}[t]
\centering
\includegraphics[width=0.5\textwidth]{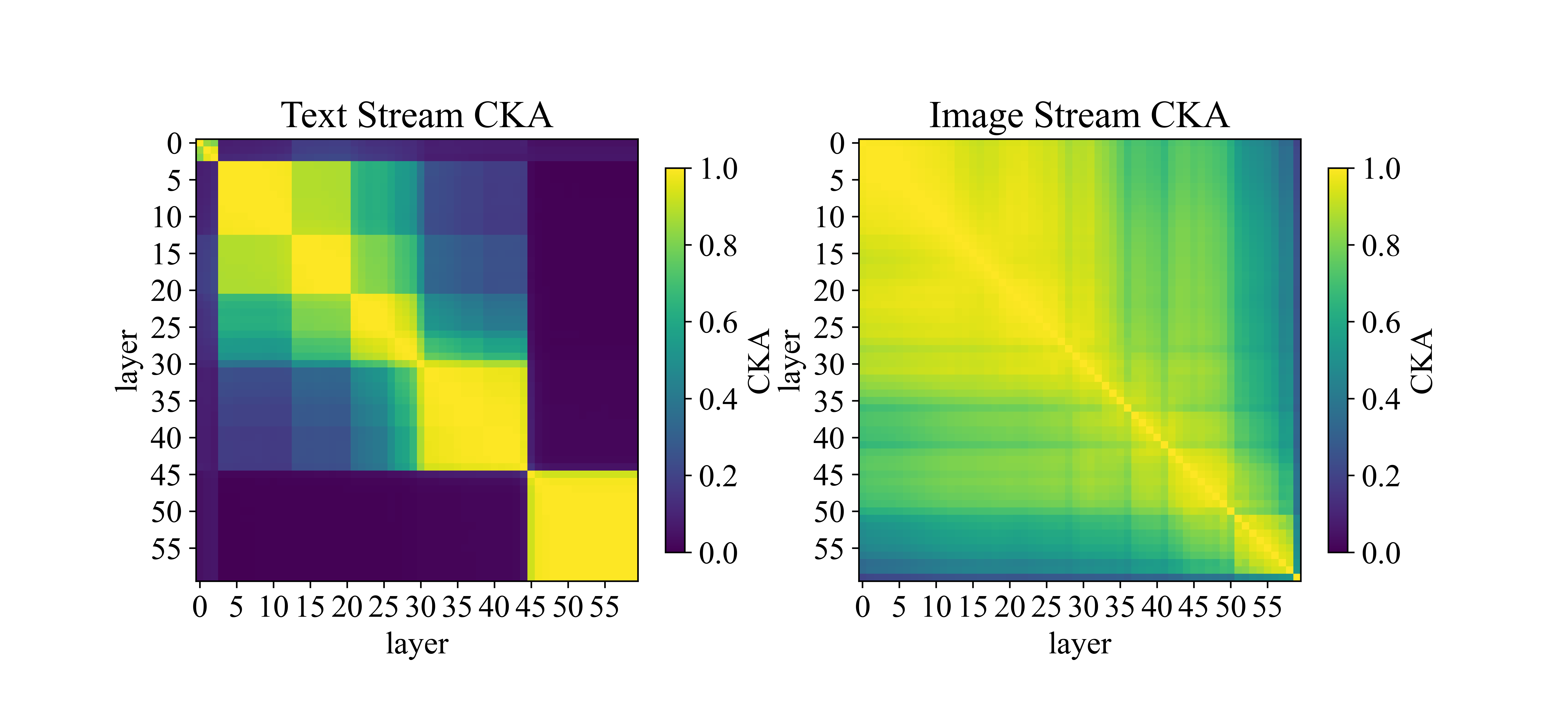}
\caption{MMDiT's text/image stream CKA heatmaps: Text stream shows high cross-layer similarity with substantial redundancy; Image stream exhibits smooth diagonal similarity decay, reflecting sequential feature evolution with minimal redundancy.}\label{stream-level CKA}
\end{figure}

MMDiT employs a dual-stream architecture to achieve fine-grained cross-modal alignment. However, as shown in Fig.\ref{stream-level CKA}, the heatmaps reveal high representational similarity across layers in the text stream, indicating functional redundancy between the text streams of certain adjacent layers and enabling compression. We define the text stream-level redundant layers as $\mathcal{R}_{txt}$. Furthermore, by measuring the similarity before and after replacing FFNs with linear projectors, we observe that the MSE loss of certain layers is extremely small. This indicates that the FFNs of these layers can be replaced with linear transformations without significantly degrading model performance, we define FFN redundant layers as $\mathcal{R}_{ffn}$, and apply width-wise pruning accordingly. This analysis provides a theoretical foundation for subsequent model pruning and optimization. For additional details on layer selection, refer to the appendix.

As shown in Stage 2 of Algorithm \ref{PPCL_algorithm}, for stream-level pruning, we replace all parameters in the text stream of the $p$-th ($p\in\mathcal{R}_{txt}$) layer except for those in QKV projectors with two lightweight linear projectors $l_p^z$ and $l_p^h$. To preserve cross-modal alignment while reducing parameter overhead, we position $l_p^z$ before the QKV projectors. Specifically, we use the intermediate text representations $z_{r}^{txt}$ and the text stream outputs $h_{r}^{txt}$ from the preceding dual-stream block at the $r$-th layer as inputs to the projectors, respectively:
\begin{equation}
z_{p}^{txt} = l^z_p(z_{r}^{txt}),\ h_p^{txt} = l_p^h(h_{r}^{txt}).
\end{equation}
Similarly, for FFN pruning, we replace the FFNs in the image and text streams of the $q$-th layer ($q\in\mathcal{R}_{ffn}$) with compact linear projectors $l_q^{img}$ and $l_q^{txt}$:
\begin{equation}
h_q^{img} = l_q^{img}(g_q^{img}),\ h_q^{txt} = l_q^{txt}(g_{q}^{txt}),
\end{equation}
where $g_q^{img}$ and $g_q^{txt}$ denote the gating outputs of the image stream and text stream. In addition to the term in Eq.\ref{depth loss}, the distillation loss incorporates a linear alignment loss. The total loss is:
\begin{equation}
\label{width total}
\mathcal{L}_{width} = \sum_{j \in \mathcal{R}_{txt} \cup \mathcal{R}_{ffn}} (\mathcal{L}_{width}^j + \mathcal{L}_{linear}^j),
\end{equation}
\begin{equation}
\label{width loss}
\mathcal{L}_{width}^j = \left\| \mathrm{Norm}(S_{width}^j(T_{j-1}^D)) - \mathrm{Norm}(T_j^D) \right\|_2^2,
\end{equation}
\begin{equation}
\label{width linear}
\mathcal{L}_{linear}^j = 
\begin{cases}
    \left\| z_{j}^{txt} - T_{j}^{z} \right\|_2^2+\left\| h_{j}^{txt} - T_{j}^{txt} \right\|_2^2, & \text{if } j \in \mathcal{R}_{txt} \\
    \left\| h_{j}^{img} - T_{j}^{img} \right\|_2^2+\left\| h_{j}^{txt} - T_{j}^{txt} \right\|_2^2, & \text{if } j \in \mathcal{R}_{ffn}
\end{cases}
\end{equation}
where $T_{j}^{z}$, $T_{j}^{img}$, and $T_{j}^{txt}$ denote the intermediate text representations, image stream outputs, and text stream outputs of the $j$-th layer of the teacher model, respectively.

\section{Experiment}

\begin{table*}[t]
\centering
\small
\setlength{\tabcolsep}{1.5pt}
\begin{tabular}{c|c|ccc|ccccccccc|c}
\hline
\multirow{2}{*}{Models} & \multirow{2}{*}{Methods} & \multirow{2}{*}{P.(B)$\downarrow$} & \multirow{2}{*}{M.(\%)$\downarrow$} & \multirow{2}{*}{L.(ms)$\downarrow$} & \multirow{2}{*}{DPG$\uparrow$} & \multirow{2}{*}{GenEval$\uparrow$} & \multicolumn{2}{c}{LongText$\uparrow$} & \multicolumn{2}{c}{OneIG$\uparrow$} & \multicolumn{3}{c|}{T2I-CompBench$\uparrow$} & \multirow{2}{*}{R.(\%)$\downarrow$} \\ \cline{8-14}
 & &  &  &  &  &  & EN & ZH & EN & ZH & B-VQA & UniDet & S-CoT &  \\ \hline
\multirow{6}{*}{FLUX.1-dev} & Base model & 12 & 100 & 715 & 83.8 & 0.665 & 0.607 & - & 0.434 & - & 0.640 & 0.426 & 78.57 & 0 \\
& FLUX.1 Lite & 8 & 78.8 & 572 & 82.1 & 0.623 & 0.643 & - & 0.378 & - & 0.547 & 0.379 & 77.21 & 6.09 \\
& Chroma1-HD & 8.9 & 82.5 & 1714 & 84.0 & 0.593 & 0.708 & - & 0.490 & - & 0.621 & 0.339 & 76.43 & 1.02 \\
& Dense2MoE & 12 & 100 & 312 & 73.6 & 0.403 & - & - & - & - & 0.473 & 0.311 & 76.25 & 21.52 \\
& TinyFusion & 8 & 74.4 & 534 & 77.2 & 0.511 & 0.502 & - & 0.345 & - & 0.584 & 0.369 & 74.17 & 13.80 \\
& HierarchicalPrune & 8 & 74.4 & 543 & 75.7 & 0.503 & 0.525 & - & 0.351 & - & 0.579 & 0.371 & 74.99 & 13.38 \\
& \textbf{PPCL}(8B) & 8 & 74.4 & 535 & 80.0 & 0.605 & 0.564 & - & 0.456 & - & 0.615 & 0.391 & 78.15 & 4.03 \\
\hline
\multirow{2}{*}{FLUX.1 Lite} & Base model & 8 & 78.8 & 572 & 82.1 & 0.623 & 0.643 & - & 0.378 & - & 0.547 & 0.379 & 77.21 & 0 \\
& \textbf{PPCL}(6.5B) & 6.5 & 69.2 & 428 & 81.2 & 0.593 & 0.620 & - & 0.371 & - & 0.581 & 0.398 & 76.91 & 0.07 \\
\hline
\multirow{6}{*}{Qwen-Image} & Base model & 20 & 100 & 2625 & 88.9 & 0.870 & 0.943 & 0.946 & 0.539 & 0.548 & 0.709 & 0.532 & 82.47 & 0 \\
& TinyFusion & 14 & 79.4 & 1789 & 80.7 & 0.739 & 0.859 & 0.857 & 0.500 & 0.498 & 0.689 & 0.464 & 78.99 & 8.75 \\
& HierarchicalPrune & 14 & 79.4 & 1786 & 83.3 & 0.766 & 0.884 & 0.881 & 0.509 & 0.496 & 0.706 & 0.487 & 79.94 & 6.49 \\
& \textbf{PPCL}(14B) & 14 & 79.4 & 1792 & 87.9 & 0.847 & 0.929 & 0.946 & 0.536 & 0.538 & 0.750 & 0.524 & 82.15 & 0.42 \\
& \textbf{PPCL}(12B) & 12 & 71.4 & 1514 & 83.6 & 0.801 & 0.893 & 0.917 & 0.517 & 0.531 & 0.733 & 0.529 & 81.94 & 3.03 \\
& \textbf{PPCL}(10B) & 10 & 66.9 & 1462 & 81.7 & 0.784 & 0.871 & 0.885 & 0.485 & 0.501 & 0.701 & 0.499 & 79.24 & 6.88 \\
& \textbf{PPCL}(Fine-tuning) & 10 & 66.9 & 1462 & 86.7 & 0.828 & 0.902 & 0.931 & 0.499 & 0.519 & 0.715 & 0.515 & 81.33 & 3.29 \\
\hline
\end{tabular}
\caption{A comprehensive comparison in terms of performance and efficiency. P., M., L. and R. denote model parameter count (Billion), GPU memory usage, inference latency (milliseconds), and average performance drop, respectively.}
\label{overall_exp}
\end{table*}

\subsection{Experimental Setup}

\textbf{Dataset and Experimental Settings.} We sample 100,000 images from LAION-2B-en. For each image, we generate refined descriptions using Qwen2.5-VL\cite{Qwen2.5-VL} and produce corresponding training pairs using Qwen-Image. We apply a three-stage strategy to obtain the pruned model: Depth-wise pruning based on redundant intervals identified via linear probing for 6k steps (8 H20 GPUs, micro-batch=2), width-wise pruning for 2k steps (identical settings), and full-parameter fine-tuning for 1k steps (micro-batch=4). Additionally, for Qwen-Image, leveraging the plug-and-play property, we obtain 12B and 14B models by replacing selected student layers in the 10B model with their corresponding teacher layers. All training runs leverage BF16 mixed-precision and gradient checkpointing to improve memory efficiency. We use the AdamW\cite{Adamw} optimizer throughout ($\beta_1$=0.9, $\beta_2$=0.95, weight decay=0.02), and fix the random seed to 42 for all inference. For more detailed hyperparameters, please refer to the appendix. 

\textbf{Comparative Methods.} We reimplement the pruning pipelines of TinyFusion and HierarchicalPrune on Qwen-Image and FLUX.1-dev, given the unavailability of their official open-source models. For TinyFusion, we adopt learnable gating parameters to identify layers for removal and apply standard distillation for restoration. For HierarchicalPrune, we first apply HPP to remove less important layers, followed by PWP that freezes critical early layers and restores performance via knowledge distillation. Since HierarchicalPrune further incorporates INT4 weight quantization in its final stage, we omit quantization in our implementation to ensure a fair comparison. We set pruning ratios to 33\% (12B$\to$8B) for FLUX.1 and 30\% (20B$\to$14B) for Qwen-Image. For Dense2MoE with FLUX.1, we report the results directly from the original paper. In addition, for comparisons on FLUX.1, we include two open-source pruned variants, Chroma1-HD and FLUX.1 Lite, and further apply our pruning strategy to FLUX.1 Lite.

\begin{figure}[h!]
\centering
\includegraphics[width=0.5\textwidth]{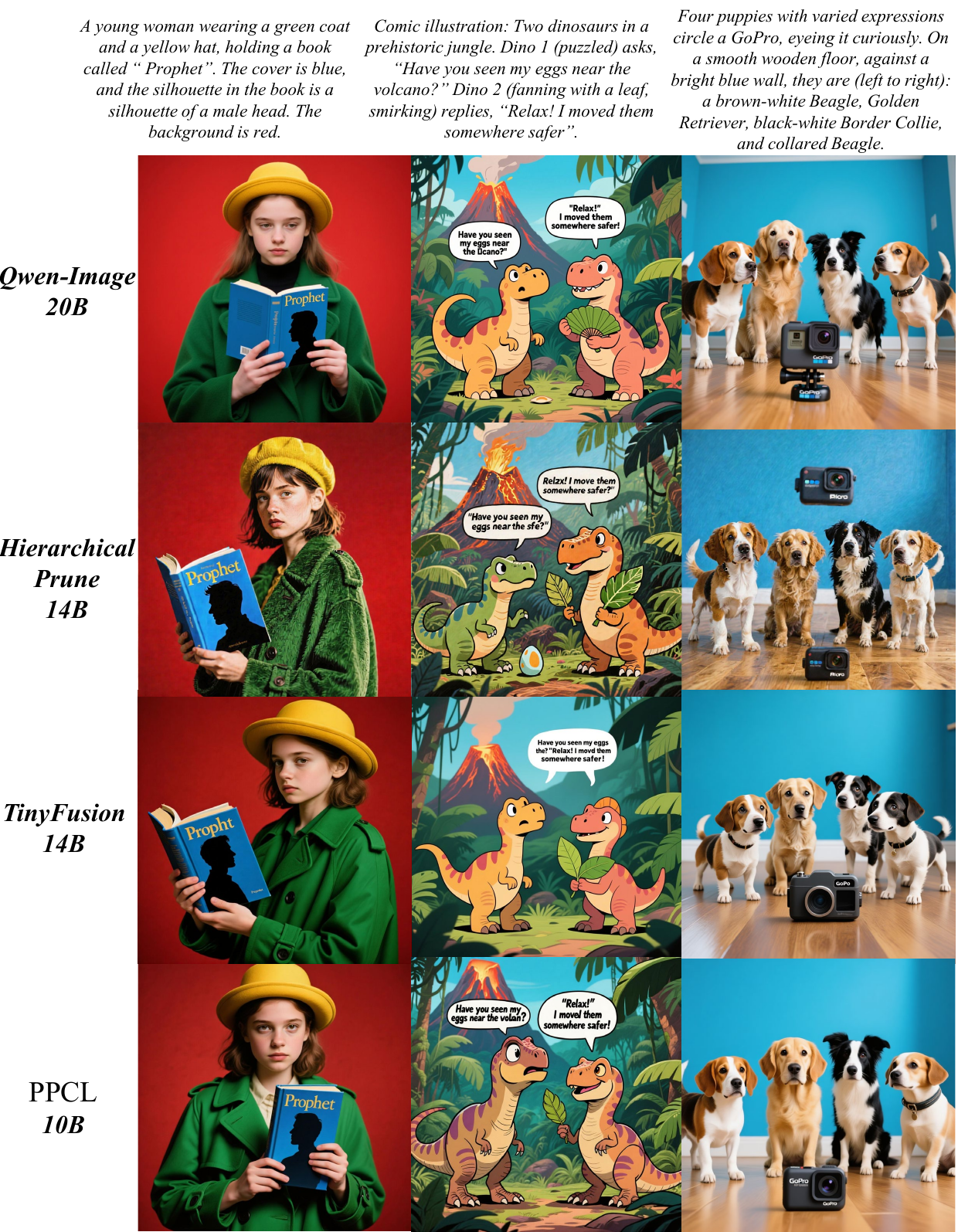}
\caption{A subjective comparison result: The first row shows the teacher model. The second and third rows display the 14B pruned HierarchicalPrune and TinyFusion models, respectively. The last row illustrates the effect of our method pruned to 10B.}\label{Effect_Comparison}
\end{figure}

\textbf{Evaluation Metrics.} To evaluate the general generation capability of the models, we conduct experiments on four public benchmarks: DPG\cite{DPG}, GenEval\cite{Geneval}, OneIG-Bench\cite{OneIG}, and T2I-CompBench\cite{T2i-compbench}. To assess long-text rendering, we use the LongText-Bench\cite{LongText} to evaluate English and Chinese long-text-to-image performance. In addition, we compare the model size, GPU memory usage, and inference latency to assess computational efficiency.

\subsection{Main Results}
\textbf{Objective results.}
As shown in Tab.\ref{overall_exp}, we organize the results into three parts. The top part of the table presents a comparison of pruning and compression experiments based on FLUX.1-dev. Among all models, Chroma1-HD achieves the lowest average performance drop (1.02\%). However, in addition to pruning its parameters down to 8.9B, it incurs a significant slowdown in inference time (1714 ms), exceeding twice that of the base model (715 ms). In the comparison of models pruned to 8B parameters, our method exhibits the smallest average performance degradation (4.03\%), outperforming both TinyFusion (13.80\%) and HierarchicalPrune (13.38\%). Additionally, we conduct extra pruning on the FLUX.1 Lite, removing 1.5B parameters and retaining 6.5B, with a performance drop of 0.07\%. This demonstrates that our method can effectively compress already-pruned models without significant quality loss.

The bottom part presents pruning and compression experiments on Qwen-Image. When pruning the student model to half the parameter count of the teacher model, we observe an average performance drop of 3.29\% across benchmarks, with inference speed nearly doubling and GPU memory consumption reduced by about 33\%. Crucially, our plug-and-play strategy enables an effective balance between the pruning ratio and performance trade-off. Leveraging the plug-and-play property, we construct 12B and 14B variants directly from the trained 10B model without additional training, based on the layer-wise loss trends and fitting status observed during depth-wise pruning. In the comparative experiments of models compressed to 14B parameters, our method still demonstrates certain advantages over TinyFusion and HierarchicalPrune, exhibiting a performance drop of 0.42\%. Notably, the results of our 10B model are obtained after undergoing final full-parameter fine-tuning, and its performance drop (3.29\%) is comparable to that of the 12B model (3.03\%). These demonstrate that our method can maintain competitive performance with significant parameter reduction. The Flux series does not support Chinese, so Chinese-related evaluations were not conducted. Detailed experimental results are provided in the appendix.

\textbf{Subjective results.}
As shown in Fig.~\ref{Effect_Comparison}, we conduct a subjective comparison among multiple pruning methods and the teacher model. We apply HierarchicalPrune and TinyFusion separately to prune the Qwen-Image model down to 14B parameters. The results generated by HierarchicalPrune contain some visual artifacts. A possible reason is that the process of identifying and removing unimportant layers using HPP is somewhat coarse. Moreover, layer importance patterns do not strictly follow the trend where importance decreases with increasing depth. As for TinyFusion, its overall semantic alignment performance is slightly insufficient. Overall, our method demonstrates certain advantages in generation fidelity, visual quality, and text-image alignment, achieving superior results. We also conduct pruning experiments on the Qwen-Image-Edit and Qwen-Image-Edit-2509 models\cite{Qwen-image}, showing consistent effectiveness of our method. For relevant details and subjective results, please refer to the appendix.

\subsection{Ablation Study}

\textbf{Linear Probing} (LP). We uniformly prune 25 layers followed by layer distillation experiments. First, we perform layer output sensitivity analysis using CKA to select the layers to be pruned, which are then used for distillation as the baseline. Second, to verify the effectiveness of the strategy for finding inflection points via first-order difference, we determine the lower bound $v$ of substitutable consecutive layers using a simple CKA similarity threshold instead of the variation trend of first-order difference, denoted as LP-a. Third, we adopt the $[u,v]$ interval determined by LP combined with first-order difference judgment as described in this paper, denoted as LP. To further demonstrate that LP can identify the maximum subset of substitutable consecutive layers, we adjust the upper bound of some consecutive intervals $[u,v]$ from $v$ to $v+1$, denoted as LP-b. This modification is restricted to specific intervals because adjusting the upper bound $v$ of other consecutive intervals would disrupt the lower bound $u$ of the subsequent consecutive interval, violating the non-overlapping and ordered nature of the interval set. As shown in Tab.\ref{ablation}, the baseline exhibits a significant performance drop, with the average score decreasing from 0.894 to 0.731, corresponding to a average decline of 18.2\%. This indicates that structured pruning based solely on importance metrics and sequential distillation, without compensation, severely impairs model capability. The variant employing LP and sequential distillation achieves an average score of 0.761. This improvement confirms that layer redundancy estimation guided by LP leads to more effective pruning. Furthermore, both LP-a and LP-b exhibit a certain degree of performance degradation compared to LP, which validates the effectiveness of the strategy based on LP combined with first-order difference measurement. The specific distillation layer intervals chosen by different strategies will be provided in the appendix.

\textbf{Depth-wise Pruning} (DP). Replacing sequential knowledge distillation with non-sequential distillation further improves the average score to 0.848, in comparison with +LP, the average metric has improved by nearly 9 points.

\textbf{Width-wise Pruning.} We first remove the text stream and replace it with trainable linear layers (WP-text), which improves the average score to 0.86. By incorporating additional trainable linear layers, the performance loss arising from partial layer alignment in DP is mitigated. This further reduces the distributional shift between the student and teacher models, enabling the proposed approach to outperform DP despite a 1B parameter reduction in the student model. Furthermore, replacing the FFN with a linear layer (WP-ffn) yields an average score of 0.85, with only a slight drop while reducing the parameter count by 1B. 

\textbf{Fine-tuning.} Full-parameter fine-tuning after combined depth-wise and width-wise pruning achieves an average score of 0.87, which is only 2.61\% below that of the original model, while reducing the parameter count from 20B to 10B, corresponding to a 50\% reduction.

\begin{table}[t]
\centering
\footnotesize
\setlength{\tabcolsep}{1pt}
\begin{tabular}{c|cccc|ccc}
\hline
Model & LongText & DPG & GenEval & Avg. & P.(B) & L.(ms) & R.(\%) \\
\hline
Original & 0.942 & 0.885 & 0.854 & 0.894 & 20 & 2625 & 0 \\
\hline
Baseline & 0.625 & 0.763 & 0.728 & 0.706 & 12 & 1514 & 18.2 \\
+LP-a & 0.664 & 0.778 & 0.712 & 0.718 & 12 & 1514 & 19.7 \\
+LP & \textbf{0.712} & \textbf{0.795} & \textbf{0.776} & \textbf{0.761} & 12 & 1514 & \textbf{14.5} \\
+LP-b & 0.678 & 0.769 & 0.731 & 0.726 & 12 & 1514 & 18.8 \\
\hline
+DP & 0.905 & 0.836 & 0.801 & 0.848 & 12 & 1514 & 5.22 \\
\hline
+WP-text & 0.915 & 0.846 & 0.819 & 0.860 & 11 & 1501 & 3.79 \\
+WP-ffn & 0.906 & 0.835 & 0.809 & 0.850 & 10 & 1462 & 4.91 \\
\hline
+Fine-tuning & \textbf{0.916} & \textbf{0.867} & \textbf{0.828} & \textbf{0.870} & \textbf{10} & \textbf{1462} & \textbf{2.61} \\
\hline
\end{tabular}
\caption{Ablation studies of linear probing, depth-wise pruning and width-wise pruning. The \textbf{Original} denotes the original Qwen-Image (60 layers). Each subsequent row adds a specific component to the configuration of the preceding row. To compute the average, we scale the DPG scores by a factor of 0.01.}
\label{ablation}
\end{table}
\section{Conclusion and Limitations}

We propose PPCL, a structured compression framework for MMDiT. Via linear probing and CKA-based representation analysis, it locates contiguous redundant layer intervals and adopts non-sequential inter-layer distillation to mitigate error propagation, enabling plug-and-play depth-wise pruning without retraining. For width-wise compression, PPCL replaces redundant stream-level and FFN components with lightweight linear projectors. Experiments show compressed models via PPCL achieve a superior balance between performance and computational cost.

This paper has two key limitations. First, identifying inflection points through the first-order difference of CKA similarity lacks rigorous theoretical foundations and is largely a successful engineering heuristic. Second, INT4 quantization produces unsatisfactory results, presumably because pruning reduces network redundancy and narrows the quantization fault-tolerant space. We will continue our research and strive to iteratively refine the proposed scheme.

{
    \small
    \bibliographystyle{ieeenat_fullname}
    \bibliography{main}
}

\clearpage
\setcounter{page}{1}
\maketitlesupplementary

\textbf{Appendix~\ref{Details Settings}} presents a more detailed experimental setup.

\textbf{Appendix~\ref{Details Results}} provides a detailed breakdown of sub-metrics for all strategies and base models across each benchmark.

\textbf{Appendix~\ref{Subjective Qwen-Image}} presents additional subjective results of Qwen-Image pruned by PPCL.


\textbf{Appendix~\ref{Subjective Edit}} presents the subjective results of Qwen-Image-Edit pruned by PPCL.

\textbf{Appendix~\ref{Limitations}} discusses the limitations of PPCL and future work.

\section{Detailed Experimental Settings}
\label{Details Settings}

\textbf{Linear Probing.} For the training of linear probes, we set the number of training steps to 2000 and use the same training set as described in Sec.4.1. We select the Chinese subset of LongTextBench as the calibration set, which consists of 160 samples. When executing Eq.3, we perform the calculation for each sample and each timestep, then average the results over all samples and timesteps to obtain the final CKA value. For the computation of CKA similarity, we first center and normalize the input features, compute and center the Gram matrices of the linear kernels, then calculate the Hilbert–Schmidt Independence Criterion (HSIC) between the Gram matrices, and finally normalize the result to obtain the CKA similarity. For Qwen-Image, the redundant intervals identified are:
{
    \footnotesize
    \begin{align}
    \mathcal{I}_{Q} = &\{[3,4],[5,7],[8,10],[11,12],[15,24],[25,27],[29,30],\nonumber\\
    &[42,43],[45,47],[48,49],[52,53],[54,55],[56,57]\}.
    \end{align}
}As shown in Algorithm \ref{LP_algorithm}, we provide a concrete algorithmic example for the simulation process based on Qwen-Image, along with detailed explanation. In addition, the redundant intervals detected by baseline, LP-a, and LP-b in the ablation study (Sec.4.3) are as follows:
{
    \footnotesize
    \begin{align}
    \mathcal{I}^{baseline}_{Q} = &\{[5,7],[8,10],[11,13],[16,18],[19,21],[22,24],[25,27],\nonumber\\
    &[38,40],[42,44],[45,47],[48,50],[52,53],[54,55],\nonumber\\
    &[56,57]\}.\\
    \mathcal{I}^{a}_{Q} = &\{[5,7],[8,10],[11,14],[16,17],[19,21],[22,24],[25,28],\nonumber\\
    &[29,30],[42,44],[45,47],[48,50],[52,54],[56,57]\}.\\
    \mathcal{I}^{b}_{Q} = &\{[3,4],[5,9],[8,10],[11,13],[15,24],[25,28],\nonumber\\
    &[29,31],[42,44]\}.
    \end{align}
}

Similarly, for FLUX.1-dev, the redundant intervals detected by the linear probing for its single-stream and dual-stream layers are as follows:
{
    \footnotesize
    \begin{align}
    \mathcal{I}_{F}^{double} = &\{[3, 5], [6, 8], [9, 11], [12, 14], [15, 16],[17,18]\},\\
    \mathcal{I}_{F}^{single} = &\{[0, 1], [2, 3], [4, 5], [6, 7], [8, 9], [10, 11], [12, 13], [18,19],\nonumber\\
    &[20,21],[22,23],[24,25],[26,27],[28, 29], [30, 31]\},
    \end{align}
}

\textbf{Width-wise Pruning.} We employ linear projectors to replace components in the text streams and FFN modules. For text stream pruning, we identify layers with CKA similarity $\ge$ 0.999 and prioritize deeper layers, thereby determining the pruned layers. For FFN pruning, we compute the similarity of output features before and after replacing each FFN with a linear projector, and select the top-3 layers with highest similarity for pruning. Each linear projector is a linear layer, maintaining the same input and output dimensions (3072) as the original components, which results in a significant reduction in parameter count compared to the original FFNs and text stream components.

\begin{algorithm}
\footnotesize
\caption{Illustrative Example of the Simulation Process on Qwen-Image}
\label{LP_algorithm}
\begin{algorithmic}[1]

    \Statex \textbf{Input:}
    \Statex \hspace{\algorithmicindent} Teacher model $\mathcal{T}$ with $M$ MMDiT blocks, calibration set $X$, trained linear probes $\mathcal{L}_p^{\prime} = \{ l_i \}_{i=1}^M$.
    \Statex \textbf{Output:}
    \Statex \hspace{\algorithmicindent} Redundant intervals of Qwen-Image $\mathcal{I}_{Q}$

    \medskip

    \State $\mathcal{I}_{Q} \gets \emptyset$, $u \gets 3$ \Comment{Assume the starting point $u=3$}
    \State \textbf{while} $u \leq M$ \textbf{do}
        \State \hspace{\algorithmicindent} ${\Delta}(u,k) \gets $
        \Statex \hspace{\algorithmicindent} \hspace{\algorithmicindent} $\text{CalculateDelta}(\mathcal{T}, \mathcal{L}_p^{\prime}, X, u, k),\forall k\in[u+1,M]$(Eq.4)
        \State \hspace{\algorithmicindent} \textbf{for} $k \in \{u+2, \ldots, M\}$ \textbf{do} (Eq.5)
            \State \hspace{\algorithmicindent} \hspace{\algorithmicindent} \textbf{if} $\Delta(u,k) > \Delta(u,k-1)$ \Comment{When $k=5$, the conditional statement triggers}
                \State \hspace{\algorithmicindent} \hspace{\algorithmicindent} \hspace{\algorithmicindent} $v \gets k-1$ \Comment{$v \gets 4$}
                \State \hspace{\algorithmicindent} \hspace{\algorithmicindent} \hspace{\algorithmicindent} \textbf{break}
            \State \hspace{\algorithmicindent} \hspace{\algorithmicindent} \textbf{end if}
        \State \hspace{\algorithmicindent} \textbf{end for}
        \State \hspace{\algorithmicindent} $\mathcal{I}_{Q} \gets \mathcal{I}_{Q} \cup \{[u, v]\}$ \Comment{The interval $[3, 4]$ is included in $\mathcal{I}_{Q}$}
        \State \hspace{\algorithmicindent} $u \gets v + 1$ \Comment{$u \gets 5$ and continue searching}
    \State \textbf{end while}

    \medskip
    
    \State \Return $\mathcal{I}_{Q}$
\end{algorithmic}
\end{algorithm}

\section{Detailed Experiment Results}
\label{Details Results}

\begin{table*}[t]
\centering
\footnotesize
\setlength{\tabcolsep}{1pt}
\begin{tabular}{c|c|ccccccc|cccccccc}
\hline
\multirow{2}{*}{Models} & \multirow{2}{*}{Methods} & \multicolumn{7}{c|}{DPG$\uparrow$} & \multicolumn{8}{c}{GenEval$\uparrow$} \\
 &  & Global & Entity & Attribute & Relation & Other & Overall & R.(\%) & \makecell{Single\\Object} & \makecell{Two\\Object} & Counting & Colors & Position & \makecell{Attribute\\Binding} & Overall & R.(\%) \\ \hline
\multirow{6}{*}{FLUX.1-dev} & Base model & 74.3 & 90.0 & 88.9 & 90.8 & 88.3 & 83.8 & 0 & 0.980 & 0.810 & 0.740 & 0.790 & 0.220 & 0.450 & 0.665 & 0\\
& FLUX.1 Lite & 88.8 & 87.9 & 87.2 & 89.8 & 88.9 & 82.1 & 2.03 & 0.987 & 0.747 & 0.600 & 0.798 & 0.160 & 0.440 & 0.623 & 6.32\\
& Chroma1-HD & 84.9 & 89.5 & 88.4 & 91.4 & 91.6 & 84.0 & -0.238 & 0.962 & 0.717 & 0.462 & 0.787 & 0.240 & 0.390 & 0.593 & 10.8\\
& TinyFusion & 80.2 & 85.6 & 83.6 & 84.7 & 86.7 & 77.2 & 7.88 & 0.950 & 0.524 & 0.450 & 0.681 & 0.181 & 0.283 & 0.511 & 23.2\\
& HierarchicalPrune & 78.6 & 84.6 & 85.3 & 83.8 & 84.6 & 75.7 & 9.67 & 0.850 & 0.490 & 0.510 & 0.690 & 0.170 & 0.310 & 0.503 & 24.4\\
& \textbf{PPCL}(8B) & 85.1 & 87.9 & 85.6 & 89.8 & 87.8 & 80.0 & 4.53 & 0.978 & 0.726 & 0.593 & 0.785 & 0.170 & 0.380 & 0.605 & 9.02\\
\hline
\multirow{2}{*}{FLUX.1 Lite} & Base model & 88.8 & 87.9 & 87.2 & 89.8 & 88.9 & 82.1 & 0 & 0.987 & 0.747 & 0.600 & 0.798 & 0.160 & 0.440 & 0.623 & 0\\
& \textbf{PPCL}(6.5B) & 87.7 & 87.6 & 86.7 & 88.7 & 86.5 & 81.2 & 1.09 & 0.968 & 0.715 & 0.530 & 0.784 & 0.140 & 0.420 & 0.593 & 4.81\\
\hline
\multirow{6}{*}{Qwen-Image} & Base model & 91.3 & 91.6 & 92.0 & 94.3 & 92.7 & 88.9 & 0 & 0.990 & 0.920 & 0.890 & 0.880 & 0.760 & 0.770 & 0.870 & 0\\
& TinyFusion & 80.3 & 87.9 & 87.6 & 89.0 & 85.9 & 80.7 & 9.22 & 0.987 & 0.869 & 0.762 & 0.819 & 0.430 & 0.570 & 0.739 & 15.1\\
& HierarchicalPrune & 82.1 & 91.6 & 88.7 & 90.8 & 84.3 & 83.3 & 6.30 & 0.975 & 0.919 & 0.812 & 0.840 & 0.430 & 0.620 & 0.766 & 12.0\\
& \textbf{PPCL}(14B) & 90.1 & 90.0 & 91.5 & 93.6 & 91.0 & 87.9 & 1.12 & 0.990 & 0.925 & 0.879 & 0.874 & 0.650 & 0.765 & 0.847 & 2.64\\
& \textbf{PPCL}(12B) & 87.7 & 88.8 & 89.9 & 92.7 & 89.2 & 83.6 & 5.96 & 0.968 & 0.900 & 0.856 & 0.852 & 0.543 & 0.690 & 0.801 & 7.93\\
& \textbf{PPCL}(10B) & 85.0 & 86.8 & 85.6 & 90.5 & 87.3 & 81.7 & 8.09 & 0.968 & 0.885 & 0.822 & 0.840 & 0.521 & 0.670 & 0.784 & 9.88\\
& \textbf{PPCL}(10B Finetune) & 89.7 & 89.0 & 89.9 & 92.8 & 89.6 & 86.7 & 2.47 & 0.975 & 0.920 & 0.850 & 0.904 & 0.560 & 0.760 & 0.828 & 4.82\\
\hline
\end{tabular}
\caption{Detailed experimental results on DPG and GenEval.}
\label{DPG and GenEval}
\end{table*}

\begin{table}[t]
\centering
\fontsize{6.7pt}{8.0pt}\selectfont
\setlength{\tabcolsep}{0.5pt}
\begin{tabular}{c|c|ccccccc}
\hline
\multirow{2}{*}{Models} & \multirow{2}{*}{Methods} & \multicolumn{7}{c}{OneIG-EN$\uparrow$}\\
 &  & Alignment & Text & Reasoning & Style & Diversity & Overall & R.(\%)\\ \hline
\multirow{6}{*}{FLUX.1-dev} & Base model & 0.786 & 0.523 & 0.253 & 0.368 & 0.238 & 0.434 & 0\\
& FLUX.1 Lite & 0.514 & 0.481 & 0.238 & 0.374 & 0.242 & 0.378 & 12.9\\
& Chroma1-HD & 0.811 & 0.696 & 0.250 & 0.362 & 0.333 & 0.490 & -12.9\\
& TinyFusion & 0.476 & 0.470 & 0.202 & 0.329 & 0.247 & 0.345 & 20.5\\
& HierarchicalPrune & 0.483 & 0.469 & 0.191 & 0.362 & 0.251 & 0.351 & 19.1\\
& \textbf{PPCL}(8B) & 0.753 & 0.631 & 0.217 & 0.361 & 0.316 & 0.456 & -5.07\\
\hline
\multirow{2}{*}{FLUX.1 Lite} & Base model & 0.514 & 0.481 & 0.238 & 0.374 & 0.242 & 0.378 & 0\\
& \textbf{PPCL}(6.5B) & 0.493 & 0.480 & 0.219 & 0.351 & 0.249 & 0.371 & 1.85\\
\hline
\multirow{6}{*}{Qwen-Image} & Base model & 0.882 & 0.891 & 0.306 & 0.418 & 0.197 & 0.539 & 0\\
& TinyFusion & 0.824 & 0.845 & 0.251 & 0.389 & 0.186 & 0.500 & 7.24\\
& HierarchicalPrune & 0.836 & 0.865 & 0.246 & 0.396 & 0.201 & 0.509 & 5.57\\
& \textbf{PPCL}(14B) & 0.880 & 0.886 & 0.295 & 0.413 & 0.205 & 0.536 & 0.556\\
& \textbf{PPCL}(12B) & 0.866 & 0.885 & 0.282 & 0.396 & 0.157 & 0.517 & 4.08\\
& \textbf{PPCL}(10B) & 0.839 & 0.860 & 0.249 & 0.359 & 0.121 & 0.485 & 10.0\\
& \textbf{PPCL}(10B Finetune) & 0.854 & 0.878 & 0.268 & 0.365 & 0.130 & 0.499 & 7.42\\
\hline
\end{tabular}
\caption{Detailed experimental results on OneIG-EN.}
\label{OneIG-EN}
\end{table}

\begin{table}[t]
\centering
\fontsize{6.7pt}{8.0pt}\selectfont
\setlength{\tabcolsep}{0.5pt}
\begin{tabular}{c|c|ccccccc}
\hline
\multirow{2}{*}{Models} & \multirow{2}{*}{Methods} & \multicolumn{7}{c}{OneIG-ZH$\uparrow$}\\
 &  & Alignment & Text & Reasoning & Style & Diversity & Overall & R.(\%)\\ \hline
\multirow{6}{*}{Qwen-Image} & Base model & 0.825 & 0.963 & 0.267 & 0.405 & 0.279 & 0.548 & 0\\
& TinyFusion & 0.785 & 0.889 & 0.224 & 0.365 & 0.228 & 0.498 & 9.12\\
& HierarchicalPrune & 0.796 & 0.895 & 0.214 & 0.349 & 0.224 & 0.496 & 9.49\\
& \textbf{PPCL}(14B) & 0.818 & 0.958 & 0.241 & 0.395 & 0.265 & 0.538 & 1.82\\
& \textbf{PPCL}(12B) & 0.805 & 0.961 & 0.270 & 0.383 & 0.219 & 0.531 & 3.10\\
& \textbf{PPCL}(10B) & 0.798 & 0.921 & 0.250 & 0.362 & 0.175 & 0.501 & 8.57\\
& \textbf{PPCL}(10B Finetune) & 0.819 & 0.937 & 0.265 & 0.376 & 0.196 & 0.519 & 5.29\\
\hline
\end{tabular}
\caption{Detailed experimental results on OneIG-ZH.}
\label{OneIG-ZH}
\end{table}

\begin{table}[t]
\centering
\fontsize{6.7pt}{8.0pt}\selectfont
\setlength{\tabcolsep}{0.5pt}
\begin{tabular}{c|c|ccccccc}
\hline
\multirow{2}{*}{Models} & \multirow{2}{*}{Methods} & \multirow{2}{*}{B-VQA$\uparrow$} & \multicolumn{3}{c}{UniDet$\uparrow$} & \multirow{2}{*}{S-CoT$\uparrow$} & \multirow{2}{*}{Overall} & \multirow{2}{*}{R.(\%)} \\ \cline{4-6}
&  &  & spatial & 3d\_spatial & numeracy &  &  & \\ \hline
\multirow{6}{*}{FLUX.1-dev} & Base model & 0.640 & 0.308 & 0.380 & 0.602 & 0.786 & 0.543 & 0\\
& FLUX.1 Lite & 0.547 & 0.292 & 0.368 & 0.574 & 0.772 & 0.510 & 6.00\\
& Chroma1-HD & 0.621 & 0.234 & 0.310 & 0.472 & 0.764 & 0.480 & 11.6\\
& TinyFusion & 0.584 & 0.246 & 0.331 & 0.530 & 0.742 & 0.486 & 10.4\\
& HierarchicalPrune & 0.579 & 0.259 & 0.320 & 0.536 & 0.750 & 0.489 & 10.0\\
& \textbf{PPCL}(8B) & 0.615 & 0.265 & 0.359 & 0.550 & 0.782 & 0.514 & 5.33\\
\hline
\multirow{2}{*}{FLUX.1 Lite} & Base model & 0.547 & 0.292 & 0.368 & 0.574 & 0.772 & 0.510 & 0\\
& \textbf{PPCL}(6.5B) & 0.581 & 0.276 & 0.363 & 0.555 & 0.769 & 0.509 & 0.352\\
\hline
\multirow{6}{*}{Qwen-Image} & Base model & 0.709 & 0.428 & 0.453 & 0.716 & 0.825 & 0.626 & 0\\
& TinyFusion & 0.689 & 0.354 & 0.375 & 0.662 & 0.790 & 0.574 & 8.34\\
& HierarchicalPrune & 0.706 & 0.386 & 0.395 & 0.6801 & 0.800 & 0.593 & 5.23\\
& \textbf{PPCL}(14B) & 0.750 & 0.412 & 0.453 & 0.707 & 0.822 & 0.629 & -0.415\\
& \textbf{PPCL}(12B) & 0.733 & 0.435 & 0.450 & 0.702 & 0.820 & 0.628 & -0.287\\
& \textbf{PPCL}(10B) & 0.701 & 0.390 & 0.429 & 0.678 & 0.792 & 0.598 & 4.50\\
& \textbf{PPCL}(10B Finetune) & 0.715 & 0.413 & 0.442 & 0.691 & 0.813 & 0.615 & 1.82\\
\hline
\end{tabular}
\caption{Detailed experimental results on T2I-CompBench. To compute the
overall score, we scale the S-CoT score by a factor of 0.01.}
\label{T2I-CompBench}
\end{table}

\textbf{DPG.} As shown in Tab.\ref{DPG and GenEval}, PPCL delivers robust performance across the dimensions of relation modeling (Relation), global semantic understanding (Global), and entity generation (Entity). Compared to three base models, it shows less than 3\% performance degradation across all these dimensions, demonstrating that the pruned model preserves strong comprehension and generation capabilities for global and relational semantics. Notably, when applied to FLUX.1-dev, PPCL even enhances performance in Global dimension. Although PPCL exhibits a slightly larger performance drop on the attribute understanding and generation (Attribute) with FLUX.1-dev compared to Chroma1-HD and FLUX.1 Lite, it still maintains a competitive advantage over TinyFusion and HierarchicalPrune. Moreover, on Qwen-Image, the performance degradation in the Attribute dimension is only 0.54\%.

\textbf{GenEval.} As shown in Tab.\ref{DPG and GenEval}, PPCL demonstrates low performance degradation in single object generation (Single Object) and colors rendering (Colors), with reductions below 0.7\%. In more challenging tasks including counting understanding (Counting), two object generation (Two Object), and position understanding (Position), most methods exhibit noticeable performance degradation on FLUX.1-dev. Nevertheless, PPCL maintains the second-highest performance across these dimensions, demonstrating its ability to maintain stable performance under complex generative instructions and effectively mitigate the degradation of key generation capabilities. Furthermore, when applied to Qwen-Image, PPCL achieves an overall performance drop of only 2.64\%, remaining highly competitive across all dimensions.

\textbf{OneIG.} As shown in Tab.\ref{OneIG-EN} and Tab.\ref{OneIG-ZH}, PPCL shows only a 1.82\% performance drop on the Chinese subset (OneIG-ZH), while on the English subset (OneIG-EN), it achieves a 5.07\% overall improvement when pruning FLUX.1-dev. PPCL performs robustly in text rendering (Text) and stylization (Style), with degradation below 2\%, demonstrating its ability to preserve core textual and stylistic features. Notably, for FLUX.1-dev, PPCL even improves text rendering performance by 20\%. In generative diversity (Diversity), PPCL shows significant gains on OneIG-EN but a slight drop on OneIG-ZH, suggesting language-dependent effects while remaining controllable. For semantic alignment (alignment) and reasoning (Reasoning), although PPCL exhibits a relatively larger performance drop on FLUX.1-dev, it still outperforms TinyFusion and HierarchicalPrune, and maintains strong stability on Qwen-Image.

\textbf{T2I-CompBench.} As shown in Tab.\ref{T2I-CompBench}, On FLUX.1-dev, PPCL achieves the lowest performance degradation of 5.33\%, outperforming all other methods. For Qwen-Image, both PPCL (14B) and PPCL (12B) surpass the base model in overall performance. PPCL demonstrates superior capabilities in visual question answering (B-VQA) and reasoning (S-CoT) dimensions, with minimal performance degradation or even improvements, indicating strong retention of complex reasoning and multimodal alignment. In spatial understanding, 3D spatial understanding, and numeracy understanding, PPCL outperforms most comparative methods on FLUX.1-dev, further validating its well-rounded and stable performance preservation. Notably, when applied to Qwen-Image, PPCL achieves nearly lossless performance in 3D spatial understanding.

\section{Additional Subjective Results on Qwen-Image}
\label{Subjective Qwen-Image}

\begin{figure*}[t]
\centering
\includegraphics[width=1\textwidth]{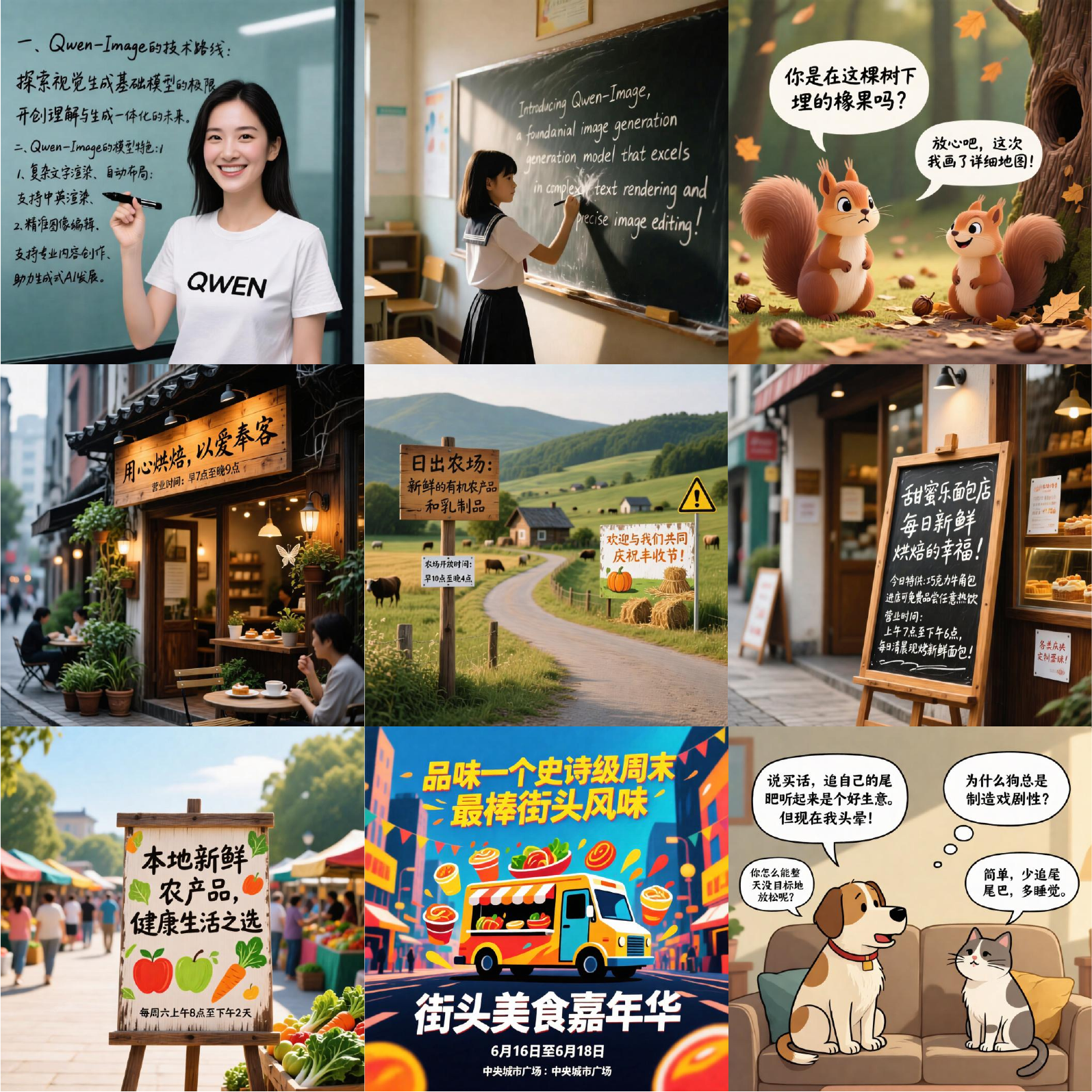}
\caption{Subjective results of PPCL (12B) with Chinese instructions.}\label{subjective result 1}
\end{figure*}

\begin{figure*}[t]
\centering
\includegraphics[width=1\textwidth]{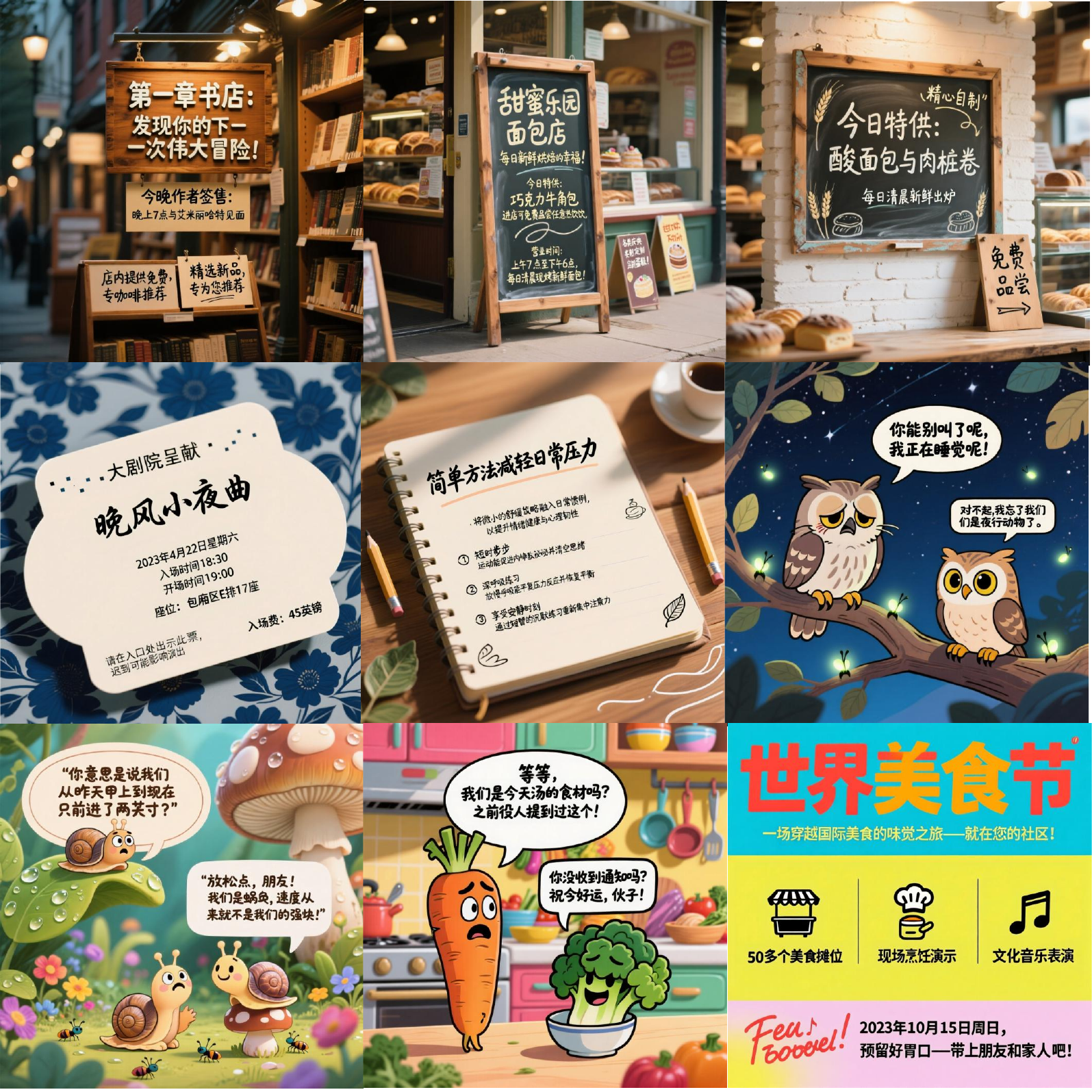}
\caption{Subjective results of PPCL (10B) with Chinese instructions.}\label{subjective result 2}
\end{figure*}

\begin{figure*}[t]
\centering
\includegraphics[width=1\textwidth]{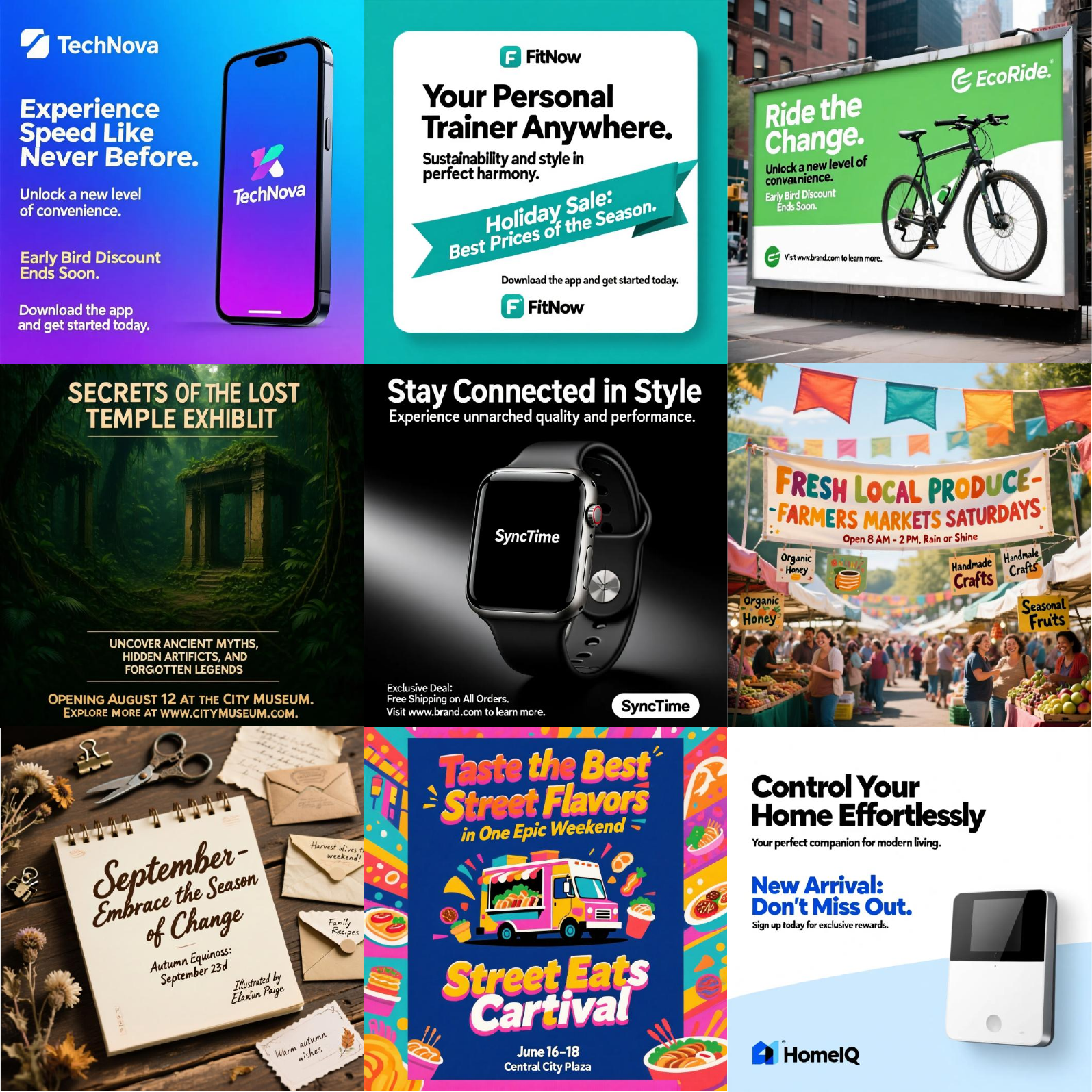}
\caption{Subjective results of PPCL (12B) with English instructions.}\label{subjective result 3}
\end{figure*}

\begin{figure*}[t]
\centering
\includegraphics[width=1\textwidth]{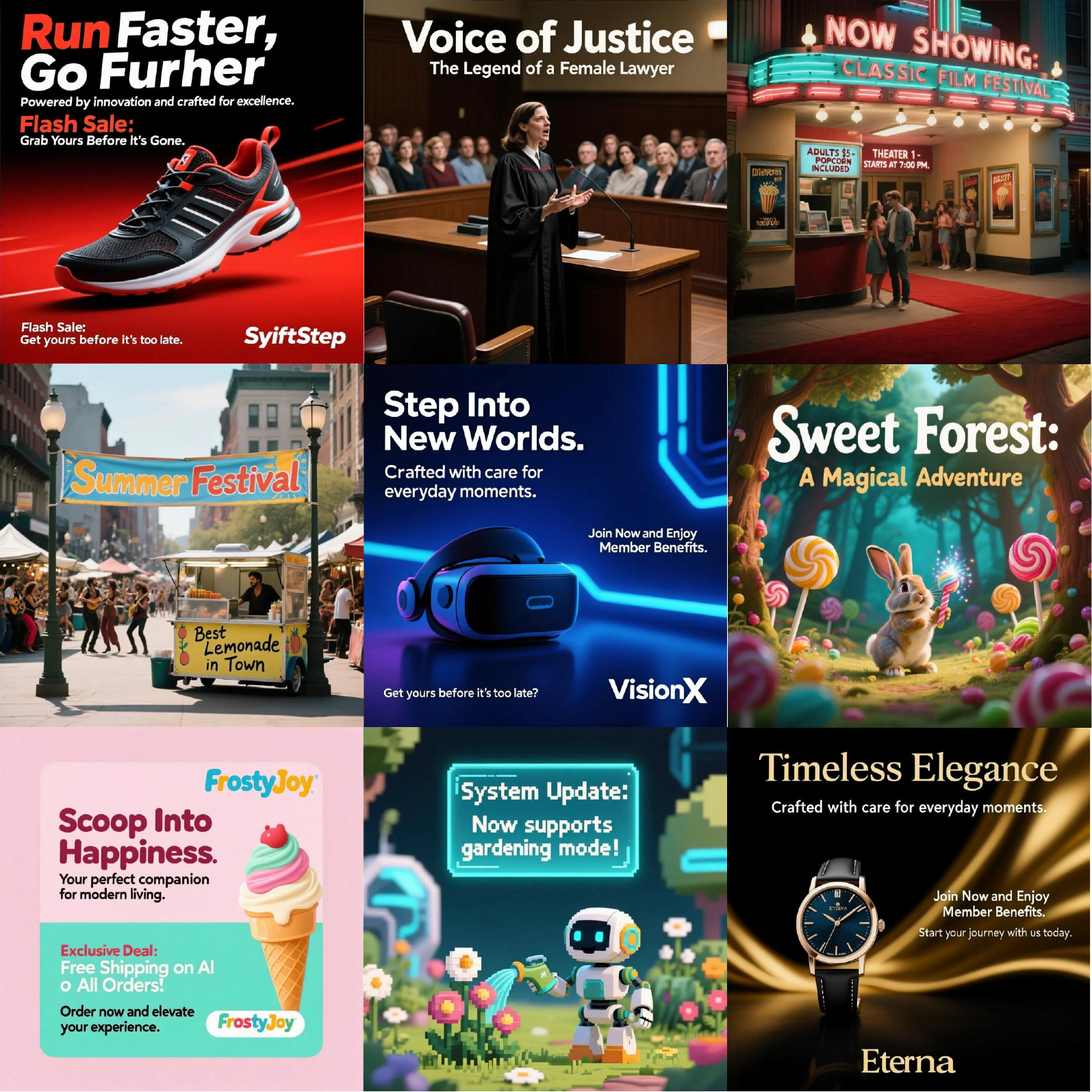}
\caption{Subjective results of PPCL (10B) with English instructions.}\label{subjective result 4}
\end{figure*}

Figs.\ref{subjective result 1}-\ref{subjective result 4} present additional subjective results of Qwen-Image pruned by PPCL. PPCL exhibits robust multilingual support, accurately rendering both Chinese and English text across diverse visual contexts while maintaining high legibility and stylistic appropriateness. It further demonstrates strong semantic visual alignment, ensuring that generated text content harmonizes seamlessly with scene atmosphere, character expressions, and overall design intent, thereby enhancing narrative coherence and contextual relevance in complex, real world applications. In summary, PPCL demonstrates sufficient capability in text generation quality, visual aesthetics, and semantic consistency to meet practical application requirements, indicating its readiness for large-scale deployment.


\section{Subjective Results on Qwen-Image-Edit}
\label{Subjective Edit}

\begin{figure*}[t]
\centering
\includegraphics[width=1\textwidth]{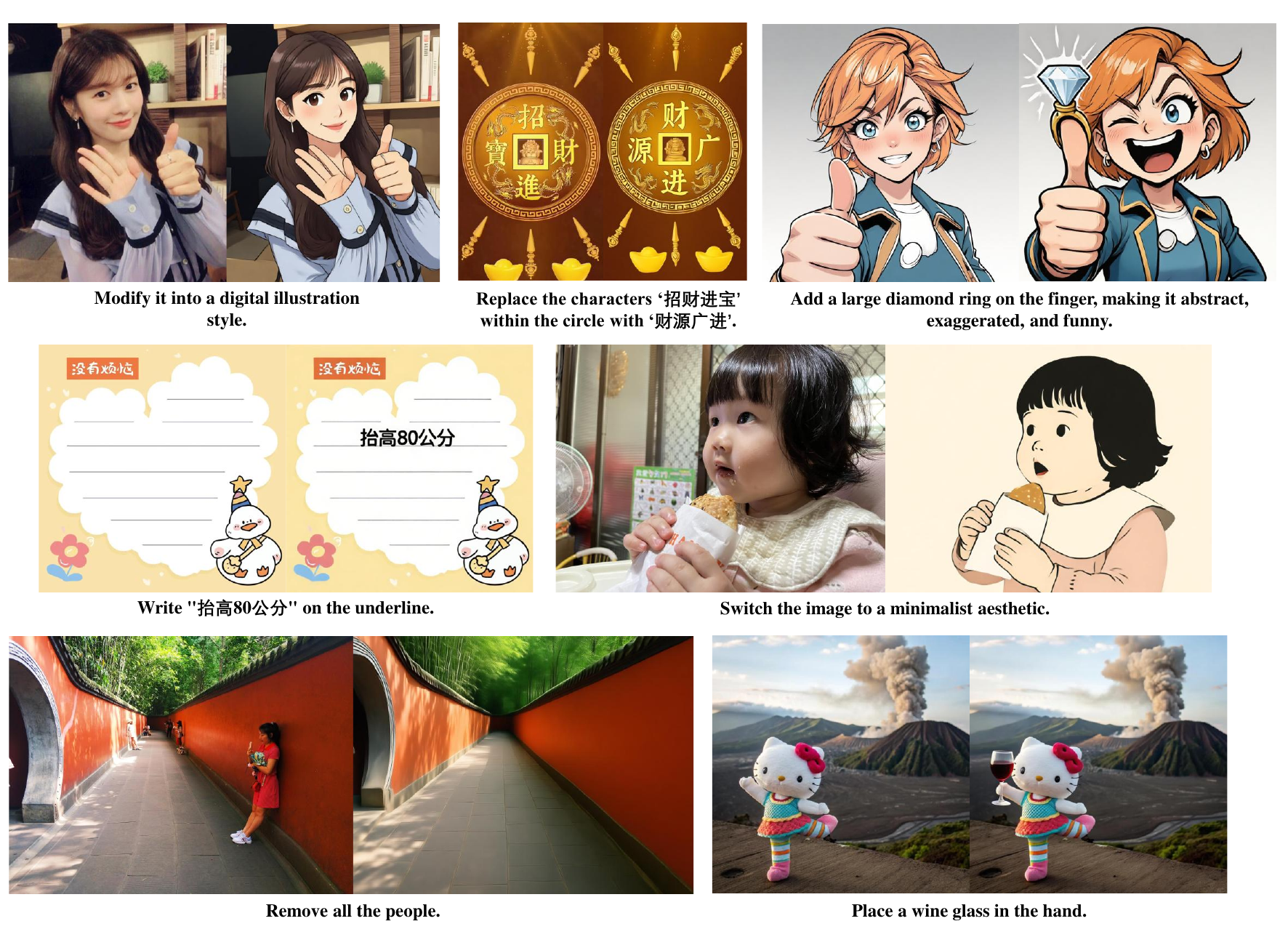}
\caption{Subjective results of the pruned Qwen-Image-Edit.}\label{subjective result 5}
\end{figure*}

\begin{figure*}[t]
\centering
\includegraphics[width=1\textwidth]{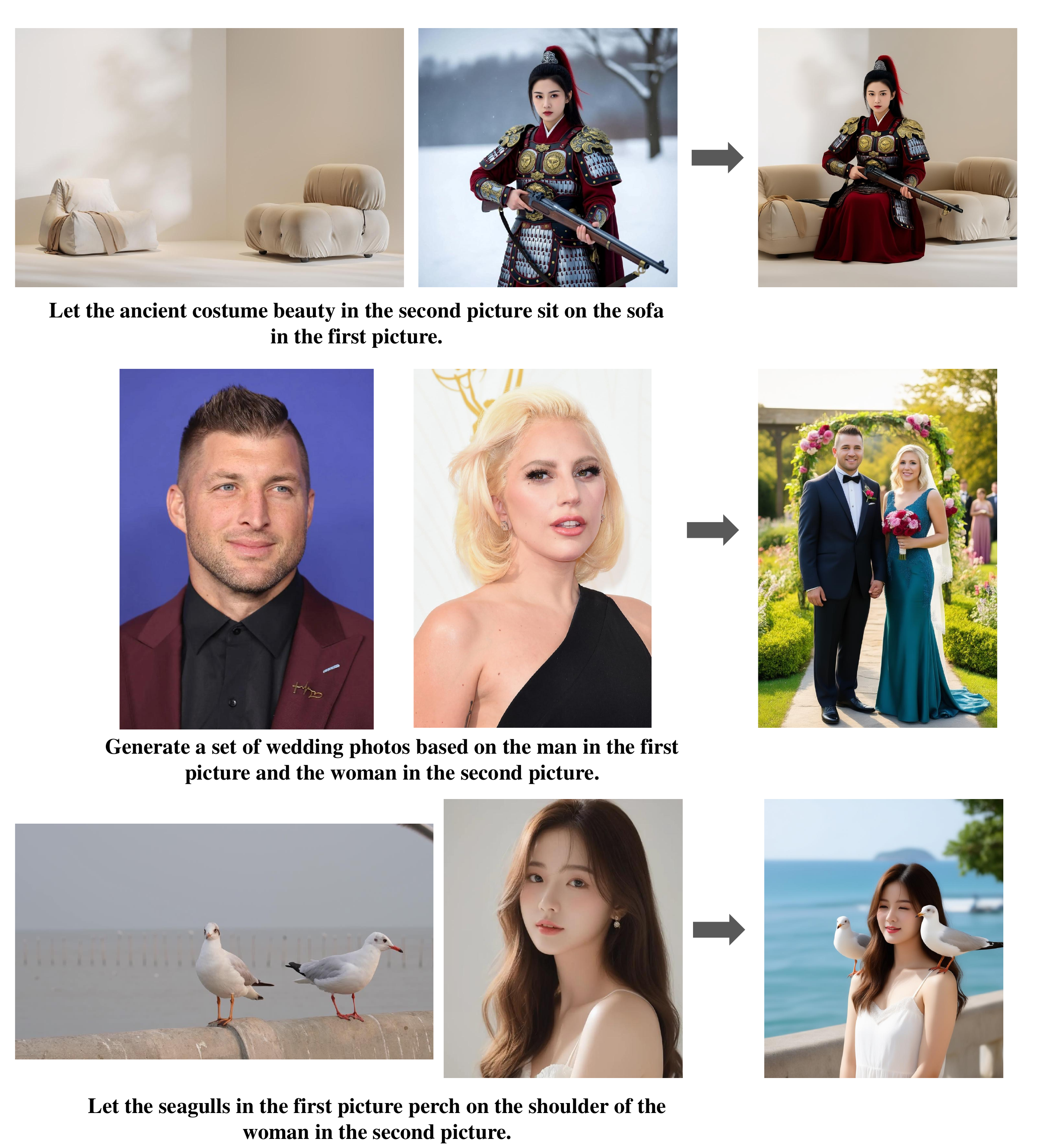}
\caption{Subjective results of the pruned Qwen-Image-Edit-2509.}\label{subjective result 6}
\end{figure*}

In addition to T2I models Qwen-Image and FLUX.1-dev, we also apply PPCL to prune image editing models, including Qwen-Image-Edit and Qwen-Image-Edit-2509. For Qwen-Image-Edit, we obtain a pruned model with 13B parameters. For Qwen-Image-Edit-2509, we obtain two pruned variants with 14B and 13B parameters, respectively. Figs.\ref{subjective result 5} and \ref{subjective result 6} present subjective results of the pruned Qwen-Image-Edit and Qwen-Image-Edit-2509, respectively.

\section{Limitations}
\label{Limitations}

\begin{figure*}[t]
\centering
\includegraphics[width=1\textwidth]{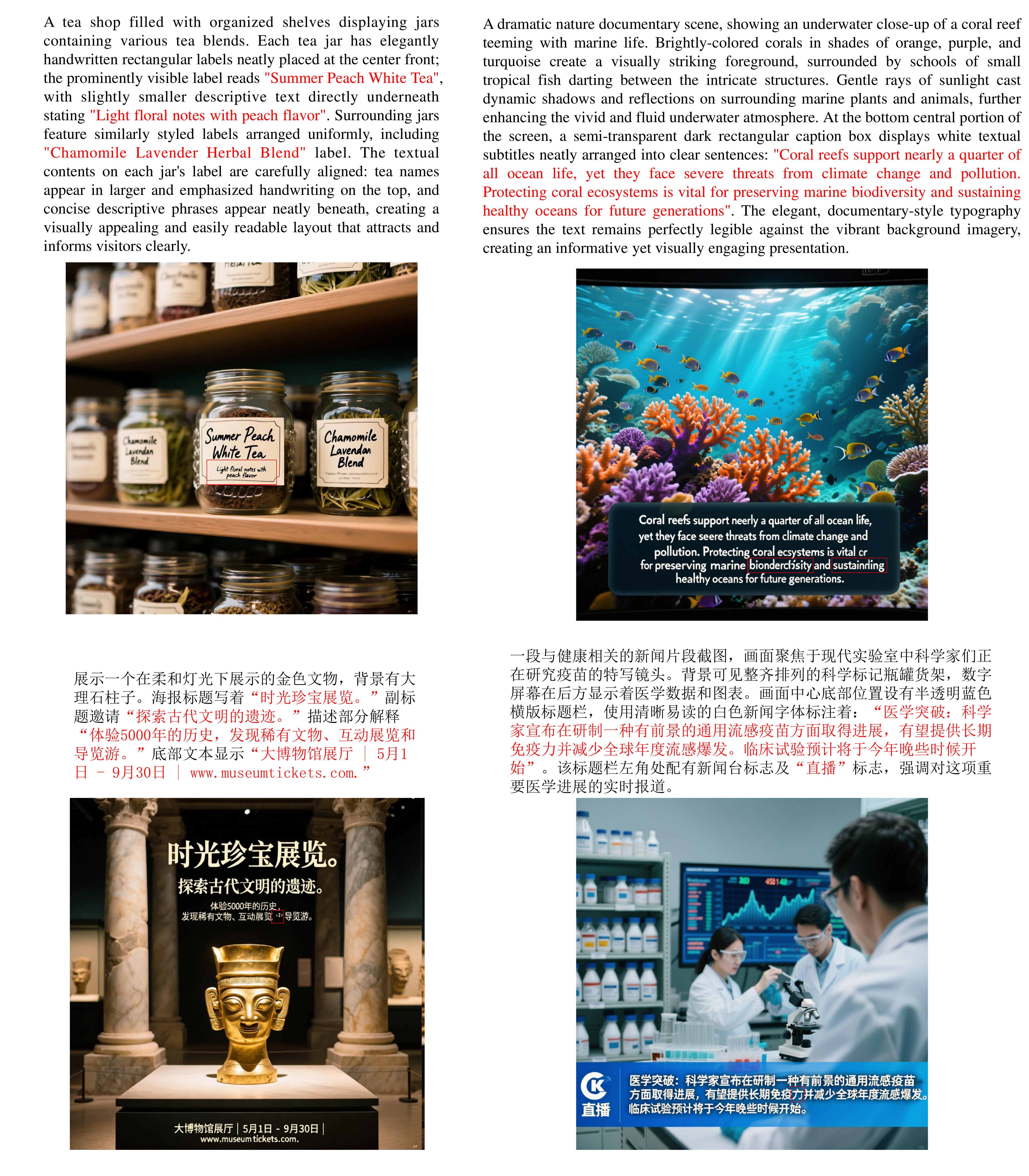}
\caption{Some failure cases.}\label{limitations}
\end{figure*}

PPCL exhibits certain limitations in rendering excessively long text and small-scale text elements. Fig.\ref{limitations} presents the corresponding failure cases. When processing extended text sequences, pruned models may struggle to maintain readability and structural integrity. In scenarios involving dense text blocks or complex layout requirements, the rendered output may exhibit inaccuracies in character representation or present issues with clarity. Similarly, for small-scale text rendering, the models sometimes fail to accurately generate fine-grained textual details, particularly in low-resolution regions or when text is placed within visually complex environments. These limitations highlight crucial directions for future improvements in the text rendering capabilities of pruned models.

Furthermore, two key limitations remain that warrant further investigation. First, the redundant intervals detection based on the first-order difference of CKA similarity lacks rigorous theoretical grounding and functions primarily as an empirical heuristic. Although effective in practice, it relies on observing local minima in CKA similarity gradients without formal proof that these points correspond to optimal pruning thresholds, leaving room for instability across diverse model architectures and datasets. Second, INT4 quantization yields suboptimal performance due to the fundamental conflict between pruning and quantization. Pruning reduces network redundancy by eliminating less critical parameters, which simultaneously narrows the quantization fault-tolerant space by concentrating parameter distributions into narrower ranges. This makes the model significantly more sensitive to quantization error, as the coarse 16-level discretization of INT4 struggles to capture the refined parameter distributions post-pruning, resulting in substantial performance degradation. Future work will focus on developing theoretically grounded strategies for detecting redundant intervals and exploring adaptive quantization strategies that account for pruning-induced structural changes to improve INT4 efficacy.

\end{document}